\pdfoutput=1
\documentclass[11pt]{article}
\usepackage[letterpaper,margin=1in]{geometry}
\usepackage{times}
\usepackage{microtype}
\usepackage{graphicx}
\usepackage{booktabs}
\usepackage{amsmath,amssymb,amsthm}
\usepackage{xcolor}
\usepackage{enumitem}
\usepackage{multirow}
\usepackage[round]{natbib}
\setcitestyle{authoryear,round,citesep={;},aysep={,},yysep={;}}
\usepackage{caption}
\usepackage{hyperref}
\definecolor{linkblue}{rgb}{0,0.08,0.45}
\hypersetup{
  colorlinks=true,
  linkcolor=linkblue,
  citecolor=linkblue,
  filecolor=linkblue,
  urlcolor=linkblue,
  pdftitle={Selective Weight Decay Opens Two Mismatches Between a Language Model and Its Embedding Table},
  pdfauthor={Issa Nakatsuka and Masayoshi Kubo}
}
\setlist{nosep,leftmargin=1.2em}

\newcommand{\iso}{\textsc{iso}}
\newcommand{\gauss}{\textsc{moment}}
\newcommand{\rows}{\textsc{rows}}
\newcommand{\bal}{b}
\newcommand{\cN}{\mbox{\textsc{neither}}}
\newcommand{\cG}{\mbox{\textsc{gains-only}}}
\newcommand{\cT}{\mbox{\textsc{table-only}}}
\newcommand{\cB}{\mbox{\textsc{both}}}
\newcommand{\fit}[1]{\resizebox{\ifdim\width>\columnwidth\columnwidth\else\width\fi}{!}{#1}}
\newcommand{\NcA}{0.84}
\newcommand{\NcB}{0.80}
\newcommand{\NcC}{0.70}
\newcommand{\DesignOut}{At a third of that output learning rate (144 further runs; Table~\ref{tab:designout}) the pattern is unchanged and the effect smaller: decaying the gains raises the moment part of the reading term by $+0.111\pm0.001$ nats with the table exempt$^{\dagger}$ and by $+0.106\pm0.022$ with the table decayed, and from resampled rows the networks of \cG{} remain faster than those of \cN{} ($-0.113\pm0.010$).}
\newcommand{\PubBest}{This comparison is at a common small step; when each checkpoint uses its best step the old ones are level with the young one over all positions and slightly behind at context positions (\S\ref{sec:limits}).}

\newcommand{\ExtraRules}{\item \emph{Failed.} ``With the trunk trained at relative step $2.6\times10^{-3}$, the reading term of the public run rises across the takeover (by at least $+0.05$ nats at context positions).'' Observed: $-0.097$ and $-0.130$.
\item \emph{Failed.} ``With the trunk trained at that step, old checkpoints are not behind from resampled rows at matched step ($\le+0.03$).'' Observed: $+0.107$ and $+0.210$.
\item \emph{Undecided.} ``With the trunk trained at that step, the step term at 300k steps is at least $+0.05$ and is reproduced by the young checkpoint within $0.06$.'' Observed: step term $+0.024$ (reproduced within $0.03$).
\item \emph{Undecided.} ``In the 7B run the step term at 200k steps is at least $+0.05$ and the replay in the young checkpoint reproduces it within $0.06$ at both ages.'' Observed: step term $+0.102$; $Z-Y=+0.082\pm0.035$ at 110k and $-0.020\pm0.043$ at 200k steps.
\item \emph{A hypothesis that was tested and rejected (not counted among the rules above).} ``With the trunk trained, the networks of \cG{} are slower than those of \cN{} even from resampled rows (so that the trunk-side lag of the public run would follow from the design).'' Observed: $-0.057\pm0.008$, faster.}
\newcommand{\IntroScope}{The public evidence is two pretraining runs of one model family; the intervention uses small networks trained on $9\times10^{7}$ tokens; the probe relearns the embedding tables for 300 updates of English text. \S\ref{sec:limits} gives the results at each checkpoint's best step and with the trunk trained, and lists what the account does not explain.}
\newcommand{\PubSevenSetting}{As a second, independent run of the same design we use OLMo-2-1124-7B (32 layers, width 4096, peak learning rate $3\times10^{-4}$), with the 30k-step checkpoint as young reference and 110k and 200k steps ($0.98$ and $0.94$ of the peak) as old ones.}
\newcommand{\PubSevenRows}{7B & 110k & $1.69$ & $+0.096$ & $+0.135$ & $+0.053$ & $+0.152$ & $-0.190$ \\
 & 200k & $2.06$ & $+0.087$ & $+0.102$ & $+0.122$ & $+0.127$ & $-0.142$ \\}
\newcommand{\PubSevenText}{\textbf{A second run.} The 7B run is a larger model of the same family with its own learning rate, batch size and data order. Between 30k and 110k steps its table grows $1.7$-fold and $\tau$ rises from $0.03$ to $0.98$ (Appendix~\ref{app:probe}). We probed all 32 layers with four probe seeds at the output learning rate $3.3\times10^{-4}$, with rules fixed in advance at the thresholds of the 1B run (Tables~\ref{tab:outlr} and~\ref{tab:seven}). The three terms are there, in a lag that does not grow between these two ages ($+0.096\pm0.046$ and $+0.087\pm0.040$ nats; $+0.26$ at context positions, $-0.06$ and $-0.08$ at first-occurrence positions). The step term is $+0.135\pm0.032$ and $+0.102\pm0.022$ nats, positive in every seed. The reading term rises by $+0.226\pm0.054$ and $+0.162\pm0.061$ at context positions (rule met), and from resampled rows at matched step the old checkpoints are ahead by $0.19$ and $0.14$ nats in every seed (rule met; at context positions alone $-0.10$ and $+0.03$). The replay in the young trunk reproduces the step term at 200k steps but not at 110k ($Z-Y=+0.082\pm0.035$, threshold $0.06$), so this rule is undecided; isotropic runs are noisier in this model.}
\newcommand{\LimRuns}{The public evidence is two pretraining runs of one model family, the second probed at one setting and with larger seed errors; checkpoint-to-checkpoint fluctuations of $0.02$ to $0.04$ nats are not contained in the seed errors, and we read the ages as a series.}
\newcommand{\AppSeven}{For OLMo-2-1124-7B, stage 1 ($4{,}194{,}304$ tokens per step, peak learning rate $3\times10^{-4}$, same decay settings), we use the checkpoints at 30k, 110k and 200k steps, where the learning rate is $0.999$, $0.982$ and $0.940$ of the peak. The rms of the table is $0.043$, $0.073$ and $0.089$; the first attention gain is $1.05\,\bal$ at 110k and $0.78\,\bal$ at 200k steps; the share was measured at ten checkpoints and crosses $1/2$ near 60k steps: it is $0.03$, $0.24$, $0.51$, $0.72$, $0.84$ and $0.98$ at 30k, 50k, 60k, 70k, 80k and 110k steps, and is $0.88$ or more afterwards. All 32 layers are used, at the output learning rate $3.3\times10^{-4}$ because the final checkpoint did not tolerate $10^{-3}$ in earlier measurements. The replay at 110k steps gives $Y=+0.053\pm0.040$ against a step term of $+0.135\pm0.032$. A repetition of the cells of probe seed 2 at 30k steps from an earlier stage of the project differs from the present runs by $0.039$ nats (\iso), $0.005$ (\gauss) and $0.001$ (\rows).}

\newcommand{\NruleAll}{60}
\newcommand{\NruleOK}{48}
\newcommand{\NruleFail}{six}
\newcommand{\NruleUndec}{six}

\newcommand{\IntroWithin}{The same holds for the growth of the lag with age inside each network: between 10{,}000 and 22{,}000 updates the step term appears only with an exempt table, the reading term rises with decayed gains, and the lag is largest in the cell of OLMo~2, where the older checkpoint is ahead once step and rows are matched.}
\newcommand{\PubOrder}{\textbf{The order of the two matchings.} Equation~\eqref{eq:ledger} matches the step first and the table second. In the opposite order, with the reading term taken at the standard step and the step term measured from resampled rows (Table~\ref{tab:order}), the step term is $+0.08$ to $+0.16$ nats and the reading term rises by $+0.13$ to $+0.17$ at the five ages, so both mismatches are there in either order$^{\dagger}$. The split between them moves by $0.07$ and $0.05$ nats at 100k and 300k steps and by $0.01$ in between; in this order the step term is clear at 100k steps as well ($+0.082\pm0.013$).}

\title{Selective Weight Decay Opens Two Mismatches Between\\ a Language Model and Its Embedding Table}
\author{%
Issa Nakatsuka\textsuperscript{1}\qquad Masayoshi Kubo\textsuperscript{2}\\[2ex]
\normalsize\textsuperscript{1}School of Informatics and Mathematical Science, Faculty of Engineering, Kyoto University\\
\normalsize\textsuperscript{2}Graduate School of Informatics, Kyoto University\\[1.5ex]
\normalsize Correspondence: \href{mailto:nakatsuka.issa.23f@st.kyoto-u.ac.jp}{\texttt{nakatsuka.issa.23f@st.kyoto-u.ac.jp}} (Issa Nakatsuka)%
}
\date{}

\begin{document}

\maketitle

\begin{abstract}
Language models become slower to relearn a new vocabulary the longer they were pretrained, a symptom that has been read as lost plasticity. We trace this lag in OLMo~2 to one choice of the optimizer: the input embedding table is exempt from weight decay, while the gains that scale the output of every sublayer are decayed. The gains settle at the equilibrium of weight decay; the table keeps growing. A new table then meets two mismatches. \emph{Scale}: a fixed learning rate is a shrinking step relative to the table. \emph{Reading}: once the table has outgrown the gains, the layers above the first read raw embedding rows, and a new table without the statistics of the old rows falls behind. A $2\times2$ intervention on pretraining (table decayed or exempt $\times$ gains decayed or not) reproduces both mismatches and their growth with age; decaying the table removes the first and limits the second. Along the public checkpoints the lag of the standard probe is the sum of the two mismatches minus an advantage of age: with the step scaled to the table and a new table made of resampled old rows, older checkpoints relearn \emph{faster} at a common relative step and as fast as young ones when each uses its best step. With the trunk frozen, what drifts with age is the interface between table and trunk rather than the trunk's ability to learn. A second public run of a larger model shows the same three terms. Decaying the table costs $0.04$ to $0.06$ nats of pretraining loss in our small networks; users of released checkpoints can instead scale the embedding learning rate with the table and start new embeddings from the statistics of the old ones. The account covers embedding relearning with the trunk frozen or trained along at a small fine-tuning rate.
\end{abstract}

\section{Introduction}

Networks that keep learning can lose the ability to learn, and weight decay is among the standard protections against this loss of plasticity \citep{dohare2024loss,lyle2023understanding,lyle2024disentangling,kumar2023maintaining}. Large language models, trained with weight decay and normalization throughout, also become harder to adapt as pretraining proceeds: late checkpoints are worse after fine-tuning \citep{springer2025overtrained}, small models trained on a stream of languages learn a held-out language less efficiently \citep{hernandezgarcia2026scale}, and relearning token embeddings for a new language is data-hungry unless the embeddings were reset during pretraining \citep{chen2023improving}. Which parameter groups to decay is an open design choice \citep{wang2026negligible}, and what that choice does to later adaptation has not been isolated by intervention.

\begin{figure*}[t]
\centering
\includegraphics[width=\textwidth]{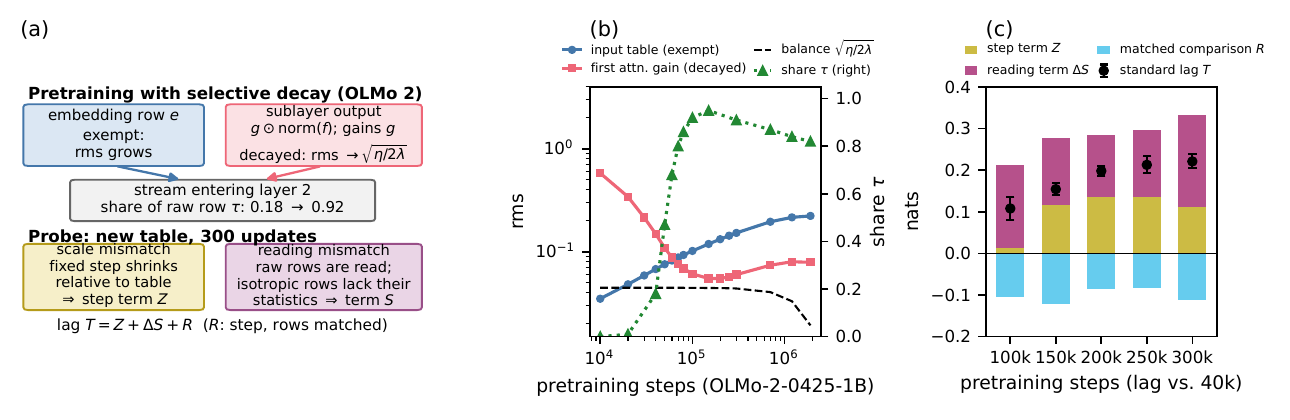}
\caption{(a) The account. OLMo~2 exempts the embedding table from weight decay and decays the gains that scale every sublayer's output; the table grows, the gains fall to the balance, and the stream that enters the second layer becomes the raw embedding row. A new table then meets a scale mismatch and a reading mismatch. (b) Public 1B run: rms of the input table (exempt), rms of the output gain of the first attention block (decayed), the balance $\bal=\sqrt{\eta/2\lambda}$ at the current learning rate, and the share $\tau$ of the raw row in the stream entering the second layer (right axis). (c) Public 1B run, frozen trunk, output learning rate $10^{-3}$: the lag $T$ of the standard relearning probe behind the 40k-step checkpoint is the sum of a step term $Z$, the rise of the reading term $\Delta S$, and a matched comparison $R$ in which old checkpoints are ahead.}
\label{fig:overview}
\end{figure*}

We study one clean instance. All token embeddings of a pretrained model are discarded and relearned, a procedure used for cross-lingual transfer \citep{artetxe2020cross,devries2021good} and as a measure of plasticity \citep{chen2023improving}. Along the public checkpoints of OLMo~2 \citep{olmo2024olmo2}, with an isotropic new table and one learning rate for all checkpoints, the loss after a fixed budget rises with the length of pretraining (Figure~\ref{fig:overview}c), in a stretch in which the learning rate of pretraining falls by less than $4\%$.

\textbf{The account.} OLMo~2 applies weight decay to every parameter except the token embeddings, which were exempted for stability \citep{olmo2024olmo2}. Under Adam with decoupled weight decay, a parameter whose updates are dominated by noise settles at an rms of $\bal=\sqrt{\eta/2\lambda}$ when decayed and performs a random walk when exempt \citep{kosson2024rotational}. So the gains that scale the output of every attention and feed-forward block fall towards $\bal$, and the table alone keeps growing (Figure~\ref{fig:overview}a,b). A new table then meets two mismatches.
\begin{enumerate}
\item \textbf{Scale mismatch.} A fixed learning rate is a smaller step relative to a larger table. The mechanism, norm growth lowering the effective learning rate, is known \citep{vanlaarhoven2017l2,lyle2024normalization,liu2026effective}; what we add is its source and size in a language model: the exempt embedding table, shown by intervention.
\item \textbf{Reading mismatch.} Once the table has outgrown the gains, the stream that enters the second layer is mostly the raw embedding row, and the trunk adapts to the statistics of these rows. An isotropic new table then loses to a table made of the old rows, reassigned at random so that no token keeps its row. This mismatch needs the gains to be decayed as well: with undecayed gains the sublayer outputs dominate the stream and the raw row is hardly read.
\end{enumerate}

\textbf{Evidence.} (i) A $2\times2$ pretraining intervention on networks of the OLMo~2 architecture, three seeds per cell (\S\ref{sec:design}). Decaying the table stops its growth at the balance, removes the step term and caps the share of the table in the stream at about a third, where the reading term is a quarter of its size in the OLMo~2 cell; decaying the gains is what opens the reading term. \IntroWithin{} (ii) Three results on the public checkpoints that could have failed (\S\ref{sec:public}). A \emph{young} checkpoint whose learning rate is divided by the growth factor of the table reproduces the step term within about $0.04$ nats at five ages. The reading term opens across the range of steps in which the table takes over the stream, and over 32 checkpoints of our own networks its moment-matched part follows the share of the table in the stream (rank correlation $+0.89$). And with step and rows matched, old checkpoints are $0.08$ to $0.12$ nats \emph{ahead} of the young one (and level with it when each uses its best step), so the standard lag is the two mismatches minus an advantage of age (Figure~\ref{fig:overview}c). (iii) The 7B run of the same family, an independent run of a larger model, shows the same three terms.

\textbf{What follows.} For the design of optimizers: decaying the table is the one-flag alternative that closes the first mismatch and limits the second, and it has a price: $0.04$ to $0.06$ nats of pretraining loss in our networks, besides the instability for which the table was exempted. For users of released checkpoints: scale the learning rate of new embeddings with the rms of the old table, and start them from the mean and covariance of the old rows \citep{hewitt2021initializing}. For the study of plasticity: a probe that re-initializes parameters at a fixed step from an isotropic start measures these two mismatches together with whatever the trunk has lost; here, with the trunk frozen, the trunk itself is not slower.

\textbf{Scope.} \IntroScope{}

\section{Setting}\label{sec:setting}

\textbf{Public models.} OLMo~2 \citep{olmo2024olmo2} is a decoder-only transformer without biases, with rotary positions, RMS normalization applied to the \emph{output} of each attention and feed-forward block, $h\leftarrow h+g\odot\widehat{f(h)}$ with a learned gain vector $g$, normalized queries and keys, and untied input and output embeddings. It is trained with AdamW ($\lambda=0.1$); the official configuration (\texttt{decay\_embeddings: false}, \texttt{decay\_norm\_and\_bias: true}) exempts the input table and decays the trunk matrices, the output table and all normalization gains. The OLMo~2 paper motivates the exemption by stability and does not discuss the decay of the gains. Our main object is the public run of OLMo-2-0425-1B (16 layers, width 2048; checkpoints between 10k and 1{,}907{,}359 steps). Its learning rate follows a cosine from $4\times10^{-4}$; it is $0.996$ of the peak at 100k steps, $0.966$ at 300k and $0.186$ at the end. We take the 40k-step checkpoint as the \emph{young} reference and 100k to 300k steps, where the learning rate is nearly constant, as the main range. \PubSevenSetting

\textbf{Own networks.} For the intervention we train networks of the same architecture class (12 layers, width 1024, a 50{,}304-token vocabulary) on English Wikipedia with the official optimizer settings except size, batch and peak learning rate ($2\times10^{-3}$; Appendix~\ref{app:probe}), up to 22{,}000 updates ($9\times10^{7}$ tokens), where the learning rate is $0.65$ of its peak. The four cells are named by which of the two groups is decayed (with $\lambda=0.1$) besides the matrices: \cN; \cG, the split of OLMo~2 (table exempt, output gains decayed); \cT{} (table decayed, gains not); and \cB. Each cell has three pretraining seeds. Where gains are decayed, these are only the gains of the two output normalizations of each layer, not those of the query/key and final normalizations, which isolates the factor (networks with the official flag: Appendix~\ref{app:arms}).

\textbf{The probe.} The input table is replaced by a new table and the output table by zeros; the trunk is frozen and both tables are trained for 300 updates of 8 sequences of 512 tokens of English Wikipedia with AdamW. The held-out loss on 32 sequences is recorded every 25 updates, and the \emph{relearning loss} $L$ is its mean over updates 250, 275 and 300. $L^{\mathrm{ctx}}$ and $L^{\mathrm{first}}$ are its parts at positions whose target token did (\emph{context} positions) or did not (\emph{first-occurrence} positions) occur earlier in the sequence. A \emph{seed} selects the training text of the probe and the random streams of the new table. The output table is trained at $10^{-3}$ unless stated. For the input table, the \emph{standard} probe uses $\eta_{\mathrm{in}}=10^{-3}$ at every checkpoint; the \emph{matched-step} probe uses $\eta_{\mathrm{in}}=s\,\sigma_e$, with $\sigma_e$ the rms of the entries of the checkpoint's own table and $s=0.0147$, the relative step of the standard probe at 40k steps. This step is small: over all positions, each checkpoint's best relative step is 3 to 27 times larger (\S\ref{sec:limits}). We keep it as the convention of this probe; \S\ref{sec:limits} repeats the main comparisons at each checkpoint's best step and with the trunk trained as well.

\textbf{New tables.} \iso: i.i.d.\ $\mathcal N(0,\sigma_e^2)$ entries. \rows: each new row is the old row of a different token, assigned at random, so the table keeps every statistic of the old one and no token keeps its row. \gauss: rows drawn from $\mathcal N(\mu,\Sigma)$ with the mean and covariance of the old rows. \rows{} and \gauss{} use a population of 50{,}000 old rows (Appendix~\ref{app:probe}) and no knowledge of which token is which.

\textbf{Quantities (Table~\ref{tab:glossary}).} For a checkpoint of age $a$ we compare with the young checkpoint $a_0$. The \emph{standard lag} $T$ is what a user of the standard probe sees. The \emph{step term} $Z$ is what scaling the step with the table removes. The \emph{reading term} $S$ is the cost of an isotropic start relative to resampled rows, a contrast within one checkpoint; we write $\Delta S=S(a)-S(a_0)$, and split $S=D+H$ into the part removed by matching two moments and the rest. The \emph{matched comparison} $R$ compares old and young checkpoints when both start from resampled rows at the same relative step. By construction
\begin{equation}\label{eq:ledger}
T(a)=Z(a)+\Delta S(a)+R(a).
\end{equation}
Equation~\eqref{eq:ledger} is an identity; the evidence lies in the independent check of each term (a replay for $Z$, the intervention for $Z$ and $S$), in the sign of $R$, and in the order of the two matchings, which we vary in \S\ref{sec:public}.
The \emph{share} $\tau$ is the fraction of the power of the stream entering the second layer that is carried by the raw embedding row, measured on held-out text. Differences are means over paired seeds; $\pm$ is the standard error over probe seeds for the public runs and over the three pretraining seeds for the intervention. Every batch of runs had a rule for reading its result, with thresholds, fixed before launch. A dagger ($^{\dagger}$) marks results read by such a rule; Appendix~\ref{app:prereg} gives the count of the rules and lists every one that failed or stayed undecided.

\begin{table}[t]
\caption{Quantities. $L(a,\cdot,\cdot)$: relearning loss of the checkpoint of age $a$ from a given new table at a given input learning rate; std: $\eta_{\mathrm{in}}=10^{-3}$; m: matched step, $\eta_{\mathrm{in}}=s\,\sigma_e(a)$.}
\label{tab:glossary}
\centering\small
\setlength{\tabcolsep}{3pt}
\fit{\begin{tabular}{@{}lll@{}}
\toprule
 & name & definition \\
\midrule
$T$ & standard lag & $L(a,\iso,\mathrm{std})-L(a_0,\iso,\mathrm{std})$ \\
$Z$ & step term & $L(a,\iso,\mathrm{std})-L(a,\iso,\mathrm{m})$ \\
$Y$ & replay in young trunk & $L(a_0,\iso,\eta\,\sigma_e(a_0)/\sigma_e(a))-L(a_0,\iso,\mathrm{std})$ \\
$S$ & reading term & $L(a,\iso,\mathrm{m})-L(a,\rows,\mathrm{m})$ \\
$D$ & \ \ moment part & $L(a,\iso,\mathrm{m})-L(a,\gauss,\mathrm{m})$ \\
$H$ & \ \ arrangement part & $L(a,\gauss,\mathrm{m})-L(a,\rows,\mathrm{m})$ \\
$R$ & matched comparison & $L(a,\rows,\mathrm{m})-L(a_0,\rows,\mathrm{std})$ \\
\bottomrule
\end{tabular}}
\end{table}

\section{What selective weight decay does to scales}\label{sec:theory}

\textbf{One update rule, two regimes.} Consider one coordinate $w$ under Adam with decoupled weight decay, $w_{t+1}=(1-\eta_t\lambda\chi)\,w_t-\eta_t u_t$, where $u_t$ is the normalized update and $\chi\in\{0,1\}$ says whether the coordinate is decayed. If $u_t$ is dominated by gradient noise and uncorrelated with $w_t$, then with $c^2=\mathbb E[u_t^2]$ summed over the correlation time of the updates,
\begin{equation}\label{eq:walk}
\mathbb E\,w_{t+1}^2=(1-\eta_t\lambda\chi)^2\,\mathbb E\,w_t^2+c^2\eta_t^2 .
\end{equation}
This is the random-walk model of \citet{kosson2024rotational}. An exempt coordinate ($\chi=0$) grows as $\sigma^2=\sigma_0^2+c^2\sum_t\eta_t^2$. A decayed one ($\chi=1$) relaxes on the time scale $1/(\eta\lambda)$ to $\sigma=c\,\bal$ with the \emph{balance} $\bal=\sqrt{\eta/2\lambda}$ \citep{kosson2024rotational,kosson2026weight}. For a coordinate that is updated at every step $c\simeq1$; rows of tokens that are absent from most batches have $c<1$ (Appendix~\ref{app:walk}). Equation~\eqref{eq:walk} holds only where the loss exerts no systematic radial force, which is an assumption to be tested for each parameter group. The scale of such a group is then set by the optimizer and not by the data \citep{velikanov2026learnable}.

\textbf{In OLMo~2.} The split of OLMo~2 puts the input table in the first regime and the gains and matrices in the second. In the public 1B run both regimes are visible (Figure~\ref{fig:overview}b). The rms of the table grows from $0.020$ at initialization to $0.068$ at 40k steps, $0.102$ at 100k and $0.153$ at 300k; its square follows $\sum_t\eta_t^2$ with $c$ between $\NcC$ and $0.85$ (Appendix~\ref{app:walk}). The trunk matrices sit at $0.80$ to $0.98$ times $\bal=0.044$ between 70k and 300k steps. The output gain of the first attention block falls from $0.58$ at 10k steps to $0.15$ at 40k and $0.055$ at 150k to 200k steps, $1.24\,\bal$; it rises again when the learning rate decays, so for the gains the balance describes the constant-rate phase only.

\textbf{The table takes over the stream.} In OLMo~2 a sublayer adds $g\odot\widehat{f(h)}$ to the stream, with rms equal to that of $g$, while the embedding row enters unnormalized. The share $\tau$ of the raw row in the stream that enters the second layer is therefore set by the ratio of the table to the first-layer gains (Figure~\ref{fig:overview}a,b). In the public 1B run $\tau$ is $0.00$ at 10k steps, $0.18$ at 40k, $0.47$ at 50k, $0.68$ at 60k, $0.80$ at 70k and $0.92$ at 100k: between 40k and 100k steps the layers above stop reading a normalized mixture and start reading the raw row. We call this the \emph{takeover}. That the stream of the early layers is essentially the embedding is the premise on which the designers of OLMo~2 exempted the table, following the stability analysis of \citet{takase2025spike}; here it is not there from the start but arrives during training, at a time set by the two regimes of Eq.~\eqref{eq:walk}.

\textbf{Two ratios.} Multiplying the table and all output gains by a common factor leaves the function of the network nearly unchanged (exactly for attention, whose queries and keys are normalized), and Adam's update of the table is invariant to the scale of the gradient. To the extent that this symmetry holds, relearning depends on the table only through two ratios, the relative step $s=\eta_{\mathrm{in}}/\sigma_e$ and the ratio $\sigma_e/g$ that sets $\tau$, and on the trunk. We use it as a heuristic, not as an identity: in the public model a common factor of two changes the held-out loss by $0.1$ to $0.5$ nats. The two mismatches of the introduction are these two ratios drifting with age: $s$ falls at fixed $\eta_{\mathrm{in}}$ because $\sigma_e$ grows, and $\sigma_e/g$ rises because the table grows while the gains shrink. The first acts at once, in any trunk; the second acts through what the trunk learns while it reads raw rows (\S\ref{sec:design}).

\section{A $2\times2$ intervention on pretraining}\label{sec:design}

A public run has one design. To test that the design causes the two mismatches we train the four cells of \S\ref{sec:setting}, compare them at 22{,}000 updates, where $\bal=0.0806$, and follow the lag inside each network from 10{,}000 to 22{,}000 updates.

\begin{figure*}[t]
\centering
\includegraphics[width=\textwidth]{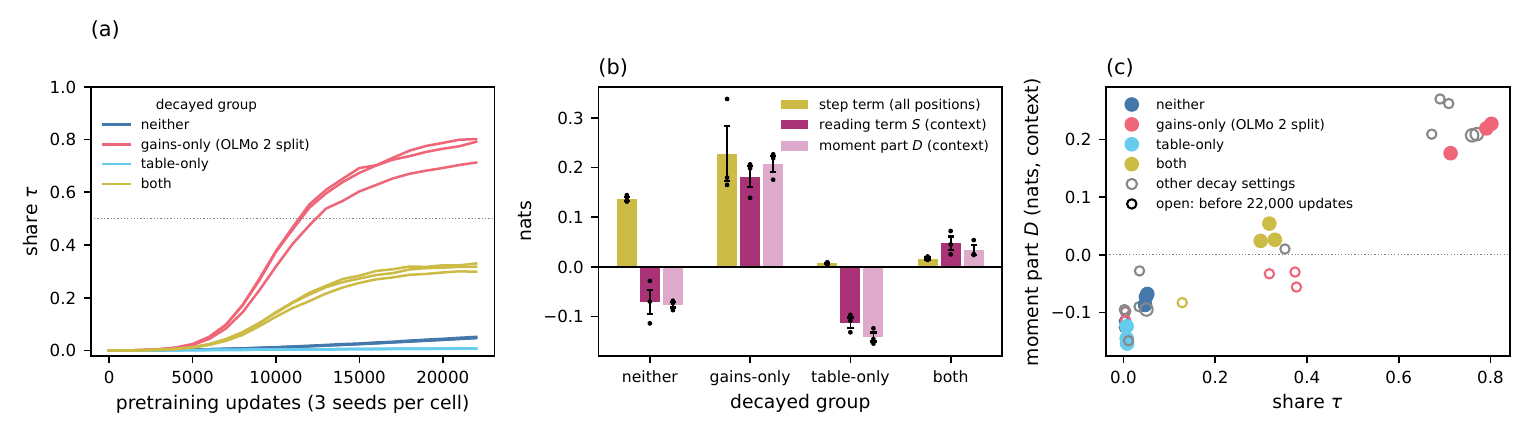}
\caption{The intervention. (a) Share $\tau$ during pretraining in the four cells, three seeds each. (b) Step term (all positions) and reading term $S$ with its moment part $D$ (context positions) at 22{,}000 updates; bars are means over the three pretraining seeds (dots), error bars their standard errors. The step term needs an exempt table, the reading term decayed gains. (c) Moment part $D$ against the share, for 32 checkpoints of 17 own networks; filled symbols are the 22{,}000-update networks of the four cells.}
\label{fig:design}
\end{figure*}

\begin{table}[t]
\caption{The four cells at 22{,}000 updates, named by the decayed group (mean over three pretraining seeds; $\pm$ standard error where it exceeds the last digit). \emph{Row}: rms of the rows of frequent tokens. \emph{Gain}: rms of the output gain of the first attention block. Both in units of $\bal=\sqrt{\eta/2\lambda}$. \emph{Growth}: row rms at 22{,}000 over 3{,}000 updates. $\tau$: share of the raw row in the stream entering the second layer. \emph{Raw}: increase of the held-out loss when the layers above the first read the mean row instead of the token's row. \emph{Loss}: held-out loss of pretraining.}
\label{tab:scales}
\centering\small
\setlength{\tabcolsep}{3.5pt}
\fit{\begin{tabular}{@{}lcccccc@{}}
\toprule
 & row$/\bal$ & growth & gain$/\bal$ & $\tau$ & raw & loss \\
\midrule
\cN & $3.00$ & $2.64$ & $14.5$ & $0.049$ & $0.01$ & $4.354$ \\
\cG & $2.57$ & $2.25$ & $1.01$ & $0.769\pm0.028$ & $3.15\pm0.14$ & $4.302$ \\
\cT & $0.96$ & $1.06$ & $14.1$ & $0.007$ & $0.00$ & $4.393$ \\
\cB & $0.98$ & $1.08$ & $0.95$ & $0.316\pm0.009$ & $1.56\pm0.10$ & $4.359$ \\
\bottomrule
\end{tabular}}
\end{table}

\textbf{Scales and the share of the table (Table~\ref{tab:scales}, Figure~\ref{fig:design}a).} A decayed table does not grow: the rows of frequent tokens stay at $0.958$ to $0.988$ times $\bal$ in the six networks with a decayed table, against $2.6$ to $3.0$ times $\bal$ and still rising in the others. The decayed output gain of the first attention block sits at $0.923$ to $1.037$ times $\bal$ whether or not the table is decayed (all layers: Appendix~\ref{app:arms}). Equation~\eqref{eq:walk} thus gives the scale of the decayed groups without a fitted coefficient. The share follows from the two scales. With the table exempt and the gains decayed, the split of OLMo~2, the growing table overtakes the gains: $\tau$ crosses $1/2$ between 11{,}400 and 12{,}400 updates and reaches $0.77$ at 22{,}000, still rising. With undecayed gains it stays at $0.05$ or below. With both decayed, table and gain meet at the same balance and $\tau$ levels off at $0.316\pm0.009$. The takeover of the stream therefore needs both halves of the split: gains that are decayed \emph{and} a table that is not. (We had predicted that $\tau$ would cross $1/2$ even with the table decayed; this failed$^{\dagger}$.) The forward dependence on raw rows follows $\tau$ (Table~\ref{tab:scales}, column ``raw'').

\begin{table}[t]
\caption{Relearning in the four cells at 22{,}000 updates (nats; mean $\pm$ standard error over three pretraining seeds, four probe seeds each). \emph{Step term}: $L(\iso)$ at the absolute learning rate that is standard for the 3{,}000-update network minus $L(\iso)$ at relative step $0.0147$, all positions. \emph{Reading term} $S$ and its \emph{moment part} $D$ (Table~\ref{tab:glossary}) at context positions. Lower block: effect of one flag at fixed other flag, paired by pretraining seed.}
\label{tab:relearn}
\centering\footnotesize
\setlength{\tabcolsep}{2pt}
\fit{\begin{tabular}{@{}lccc@{}}
\toprule
 & step term & $S^{\mathrm{ctx}}$ & $D^{\mathrm{ctx}}$ \\
\midrule
\cN & $+0.137{\pm}0.004$ & $-0.070{\pm}0.025$ & $-0.076{\pm}0.006$ \\
\cG & $+0.228{\pm}0.055$ & $+0.181{\pm}0.021$ & $+0.207{\pm}0.016$ \\
\cT & $+0.007{\pm}0.002$ & $-0.112{\pm}0.010$ & $-0.141{\pm}0.009$ \\
\cB & $+0.017{\pm}0.002$ & $+0.048{\pm}0.014$ & $+0.035{\pm}0.010$ \\
\midrule
\multicolumn{4}{@{}l}{\emph{decaying the table}}\\
\ \cT$-$\cN & $-0.130{\pm}0.004$ & $-0.042{\pm}0.015$ & $-0.065{\pm}0.015$ \\
\ \cB$-$\cG & $-0.211{\pm}0.056$ & $-0.134{\pm}0.011$ & $-0.172{\pm}0.012$ \\
\multicolumn{4}{@{}l}{\emph{decaying the gains}}\\
\ \cG$-$\cN & $+0.090{\pm}0.058$ & $+0.252{\pm}0.010$ & $+0.283{\pm}0.010$ \\
\ \cB$-$\cT & $+0.010{\pm}0.004$ & $+0.160{\pm}0.005$ & $+0.176{\pm}0.018$ \\
\bottomrule
\end{tabular}}
\end{table}

\textbf{Decaying the table removes the scale mismatch (Table~\ref{tab:relearn}, Figure~\ref{fig:design}b).} With the table exempt, a learning rate that is standard for a 3{,}000-update network is too small for the grown table at 22{,}000 updates: the step term is $+0.137$ nats in \cN{} and $+0.228$ in \cG. With the table decayed it is $+0.007$ and $+0.017$. The scale mismatch needs no takeover of the stream: it is present in \cN, whose share stays at $0.05$.

\textbf{The reading mismatch needs decayed gains, and decaying the table limits it.} The reading term is negative in the two cells with undecayed gains (an isotropic table is, if anything, the better start at context positions) and positive with decayed gains: decaying the gains raises it by $+0.252\pm0.010$ nats when the table is exempt and by $+0.160\pm0.005$ when it is decayed. Decaying the table lowers it, in the setting of OLMo~2 from $+0.181$ to $+0.048$: where the share stops at a third, an isotropic table is nearly as good a start as resampled rows. In these networks matching two moments recovers all of the reading term; the part it does not recover stays within $0.03$ nats of zero in every cell. The term has two sides. At context positions the networks of \cG{} are \emph{slower} than those of \cN{} from an isotropic table, by $+0.147\pm0.013$ nats, and \emph{faster} from resampled rows, by $-0.104\pm0.016$: decaying the gains makes the trunk a specialist of its own kind of table, not a slower learner.

\textbf{The lag grows with age inside the OLMo~2 cell (Table~\ref{tab:withinmain}).} The comparisons above are between cells of equal age. To see a lag \emph{grow} we apply Eq.~\eqref{eq:ledger} inside each network, between its checkpoints at 10{,}000 and 22{,}000 updates (held-out loss about $4.6$ and $4.3$). Over this stretch an exempt table grows $1.37$-fold and a decayed one shrinks to $0.89$ to $0.92$ of its size, because the balance falls with the learning rate. Three rules were fixed for this experiment and met$^{\dagger}$. The step term is positive in all six networks with an exempt table ($+0.035$ to $+0.139$ nats) and slightly negative with a decayed one. The reading term rises where the gains are decayed, by $+0.211\pm0.037$ and $+0.147\pm0.019$ nats at context positions, against $+0.04$ where they are not. The standard lag is largest in \cG{} ($+0.097\pm0.003$; $+0.209\pm0.012$ at context positions), and there it has the composition that we find in the public run (\S\ref{sec:public}): a step term, a rise of the reading term, and a matched comparison in which the \emph{older} checkpoint is ahead ($-0.106\pm0.002$). In \cB{} the reading term rises as well, while the share climbs to its ceiling of a third; what the decayed table limits is its level, which at 22{,}000 updates is $+0.05$ nats, a quarter of the level in \cG{} (Table~\ref{tab:relearn}). All cells share one learning-rate schedule, which falls from $0.93$ to $0.65$ of its peak over this stretch, so we read differences between cells.

\textbf{The reading term follows the share (Figure~\ref{fig:design}c).} We measured the moment part of the reading term at 32 checkpoints of 17 networks: the twelve networks above at up to three ages, and five further networks with one seed each, four of them with other decay settings (Appendix~\ref{app:arms}). However the share was reached, by which flag, at which age, with or without a growing table, $D^{\mathrm{ctx}}$ is $-0.10$ on average for $\tau\le0.06$ (16 points), $0.00$ for $\tau$ between $0.29$ and $0.40$ (7 points) and $+0.22$ for $\tau$ between $0.60$ and $0.85$ (8 points; one further point: $-0.08$ at $\tau=0.13$); the rank correlation is $+0.89$. Six of these points were measured after the band they would fall in had been fixed$^{\dagger}$, and all six fell in it.

\textbf{The one-flag alternative and its price.} Decaying the table closes the first mismatch and limits the second, but the split of OLMo~2 is the cell with the lowest pretraining loss (Table~\ref{tab:scales}; standard errors at most $0.005$). Exempting the table lowers the loss by $0.04$ to $0.06$ nats, in the direction reported by its designers, who also found training with decayed embeddings less stable \citep{olmo2024olmo2,takase2025spike}; and under decay the rows of tokens that never occur shrink to zero (Appendix~\ref{app:arms}), which can stall later training \citep{liu2026nearzero}. Decaying the gains lowers the loss by $0.03$ to $0.05$ nats in these short runs. The design that is best for the pretraining loss is thus the one that opens both mismatches. This trade is measured in small networks only.

\textbf{Conditions.} These comparisons use the relative step of the standard probe, an output learning rate of $10^{-3}$ and a frozen trunk. \DesignOut{} \S\ref{sec:limits} gives the comparison at the best step of each network and with the trunk trained.

\begin{table}[t]
\caption{The lag inside each network: Eq.~\eqref{eq:ledger} between the checkpoints at 10{,}000 and 22{,}000 updates of the same network (frozen trunk; nats; mean $\pm$ standard error over three pretraining seeds, four probe seeds each; all positions, last column context positions). $\sigma_e$ ratio: rms of the table at 22{,}000 over 10{,}000 updates. By position type: Table~\ref{tab:within}.}
\label{tab:withinmain}
\centering\small
\setlength{\tabcolsep}{2pt}
\fit{\begin{tabular}{@{}lcccccc@{}}
\toprule
cell & $\sigma_e$ ratio & $T$ & $Z$ & $\Delta S$ & $R$ & $\Delta S^{\mathrm{ctx}}$ \\
\midrule
\cN & $1.37$ & $+0.053{\pm}0.024$ & $+0.037{\pm}0.002$ & $+0.032{\pm}0.010$ & $-0.016{\pm}0.014$ & $+0.043{\pm}0.015$ \\
\cG & $1.38$ & $+0.097{\pm}0.003$ & $+0.084{\pm}0.028$ & $+0.119{\pm}0.024$ & $-0.106{\pm}0.002$ & $+0.211{\pm}0.037$ \\
\cT & $0.89$ & $+0.023{\pm}0.016$ & $-0.011{\pm}0.003$ & $+0.030{\pm}0.010$ & $+0.003{\pm}0.009$ & $+0.039{\pm}0.012$ \\
\cB & $0.92$ & $+0.036{\pm}0.016$ & $-0.014{\pm}0.002$ & $+0.100{\pm}0.016$ & $-0.051{\pm}0.004$ & $+0.147{\pm}0.019$ \\
\bottomrule
\end{tabular}
}
\end{table}

\section{The two mismatches in public runs}\label{sec:public}

\begin{table}[t]
\caption{Public 1B run, 100k to 300k steps (learning rate of pretraining within $4\%$ of its peak; all positions; nats; mean $\pm$ standard error over four probe seeds). $\sigma_e$: rms of the table relative to 40k steps. Columns as in Table~\ref{tab:glossary}; $\Delta S$ is relative to $S(40\mathrm{k})=+0.039$. $T=Z+\Delta S+R$ (Eq.~\ref{eq:ledger}).}
\label{tab:ledger}
\centering\small
\setlength{\tabcolsep}{3pt}
\fit{\begin{tabular}{@{}lcccccc@{}}
\toprule
steps & $\sigma_e$ & $T$ & $Z$ & $Y$ & $\Delta S$ & $R$ \\
\midrule
100k & $1.50$ & $+0.108{\pm}0.027$ & $+0.013$ & $+0.055$ & $+0.199$ & $-0.105{\pm}0.005$ \\
150k & $1.75$ & $+0.154{\pm}0.015$ & $+0.116$ & $+0.080$ & $+0.160$ & $-0.122{\pm}0.010$ \\
200k & $1.95$ & $+0.198{\pm}0.012$ & $+0.136$ & $+0.098$ & $+0.149$ & $-0.087{\pm}0.012$ \\
250k & $2.11$ & $+0.213{\pm}0.020$ & $+0.135$ & $+0.109$ & $+0.160$ & $-0.082{\pm}0.012$ \\
300k & $2.25$ & $+0.221{\pm}0.017$ & $+0.112$ & $+0.119$ & $+0.220$ & $-0.111{\pm}0.010$ \\
\bottomrule
\end{tabular}}
\end{table}

\textbf{The lag grows at constant learning rate.} Between 100k and 300k steps of the 1B run the standard lag $T$ rises from $+0.108$ to $+0.221$ nats in the order of age (Table~\ref{tab:ledger}, Figure~\ref{fig:overview}c; the same holds with eight probe seeds at 100k and 300k steps, $+0.091\pm0.020$ and $+0.206\pm0.012$). The learning rate of pretraining changes by $3\%$ in this range, so the rise is not an effect of learning-rate decay. At context positions $T$ is $+0.22$ to $+0.39$; at first-occurrence positions it is $+0.00$ to $+0.08$.

\textbf{Scale mismatch.} The step term $Z$, which is removed by matching the relative step, is $+0.11$ to $+0.14$ nats from 150k steps on; at 100k steps, where the table has grown $1.5$-fold, it is $+0.013\pm0.016$ and not yet separable from zero. If it is caused by the size of the table and by nothing else in the old trunk, a young trunk must show it when its own learning rate is lowered by the same factor. It does: the lag $Y$ of the 40k-step checkpoint with its learning rate divided by $1.50$ to $2.25$ is $+0.055$ to $+0.119$ nats, and $Z-Y$ is $-0.041$, $+0.036$, $+0.037$, $+0.027$ and $-0.007$ at the five ages. At the three ages not seen before, $|Z-Y|$ stays below $0.06^{\dagger}$, also on a second evaluation set of 128 sequences ($+0.028$, $+0.023$, $+0.021$). The agreement is not exact: $Z-Y$ is positive at three consecutive ages.

\textbf{Reading mismatch.} The reading term $S$ is $+0.039$ nats at 40k steps and $+0.19$ to $+0.26$ at all five later ages; its rise $\Delta S$ ($+0.15$ to $+0.22$) is of the size of the standard lag from 150k steps on and larger than it at 100k. Most of it is at context positions ($S^{\mathrm{ctx}}=+0.149$ at 40k steps and $+0.38$ to $+0.54$ later). Four additional checkpoints across the takeover show when it opens (Table~\ref{tab:switch}): $S^{\mathrm{ctx}}$ rises from $+0.15$ nats at 40k steps to $+0.44$ at 100k, in the range of steps in which the share rises from $0.18$ to $0.92$. Matching two moments recovers part of it: across the five ages the moment part $D$ rises by $+0.09$ to $+0.14$ nats over all positions, between $0.48$ and $0.86$ of $\Delta S$. The rest is the arrangement part $H$, which also opens across the takeover (Table~\ref{tab:switch}): the public trunk comes to depend on the arrangement of its rows beyond their mean and covariance. The own networks do not show this second part ($|H^{\mathrm{ctx}}|\le0.03$), so the intervention accounts for the moment part of the reading mismatch and leaves the arrangement part of the public runs as an observation.

\textbf{With step and rows matched, old checkpoints are faster.} The last column of Table~\ref{tab:ledger} compares old and young checkpoints when both start from resampled rows at the same relative step: the old checkpoints are \emph{ahead} at every age, by $0.08$ to $0.12$ nats, and behind at neither position type ($-0.01$ to $-0.11$ at context positions, $-0.13$ to $-0.16$ at first-occurrence positions). With matched moments instead of resampled rows the sum is near zero ($-0.100$ to $+0.004$), and old checkpoints are behind at context positions at four of the five ages ($+0.06$ to $+0.16$): two moments do not carry all that the trunk reads. What the standard probe shows as a growing inability to relearn is therefore the sum of a step that was not scaled with the table and of a start that lacks what the trunk has come to read, on top of a trunk that by itself relearns faster with age. \PubBest

\PubOrder

\textbf{The output learning rate.} The output learning rate of the probe, $10^{-3}$, is a convention, and a third of it gives lower losses for old checkpoints. We therefore repeated the decomposition at $3.3\times10^{-4}$ (Table~\ref{tab:outlr}). Both mismatches are there: the step term again equals the replay in the young checkpoint, and the reading term rises by about half its rise at $10^{-3}$ (at context positions by $+0.090\pm0.024$ and $+0.141\pm0.028$, both above $+0.05^{\dagger}$). What changes most is the matched comparison, so that the standard lag is only $+0.024$ and $+0.071$. The size of the standard lag depends on this convention; the presence of the two mismatches does not.

\begin{table}[t]
\caption{Public runs at an output learning rate of $3.3\times10^{-4}$ (all positions; nats; four probe seeds). Columns as in Table~\ref{tab:ledger}; the young reference is 40k steps for the 1B run and 30k steps for the 7B run.}
\label{tab:outlr}
\centering\small
\setlength{\tabcolsep}{3pt}
\fit{\begin{tabular}{@{}llcccccc@{}}
\toprule
run & steps & $\sigma_e$ & $T$ & $Z$ & $Y$ & $\Delta S$ & $R$ \\
\midrule
1B & 100k & $1.50$ & $+0.024$ & $+0.055$ & $+0.058$ & $+0.095$ & $-0.125$ \\
 & 300k & $2.25$ & $+0.071$ & $+0.154$ & $+0.127$ & $+0.093$ & $-0.176$ \\
\PubSevenRows
\bottomrule
\end{tabular}}
\end{table}

\PubSevenText

\section{Conditions and limits}\label{sec:limits}

\textbf{Best step for each checkpoint.} A lag measured at one learning rate may vanish when the rate is tuned. In the public 1B run we therefore searched grids of relative steps between $0.005$ and $0.8$ at 40k, 100k and 300k steps (two probe seeds; Table~\ref{tab:grid}; the grid was extended twice until the optima over all positions and at context positions lay inside it). With resampled rows the best losses of the old checkpoints are level with that of the young one over all positions ($+0.004$ and $+0.008$ nats; both below $+0.04^{\dagger}$). By position type they are slightly behind at context positions ($+0.051$ and $+0.030^{\dagger}$), where old checkpoints prefer small steps, and level at first-occurrence positions ($-0.009$, $+0.023$). The advantage of age seen at the common small step is thus gone at the best step, because the young checkpoint profits more from a larger step. With an isotropic table old checkpoints remain behind at their best step, by $+0.111$ and $+0.191$ nats at context positions (both above $+0.05^{\dagger}$), so the reading mismatch is not an artifact of the small step. In the intervention (Table~\ref{tab:best}) the best input learning rate is set by the gains: in eleven of the twelve networks it is $0.07$ to $0.12$ times the rms of the stream entering the second layer, $\sigma_e/\sqrt\tau$. At the best step of each network an isotropic table is still behind at context positions in \cG{} relative to \cN, by $+0.072$, $+0.078$ and $+0.083$ nats in the three seed pairs, half of the difference at the small step, and nothing remains in \cB{} relative to \cT{} ($+0.001$, $+0.000$, $-0.076$). Three rules for these comparisons were not met as stated$^{\dagger}$ (Appendix~\ref{app:prereg}).

\textbf{Training the trunk.} The probe freezes the trunk. We repeated the main comparisons with all trunk parameters trained along with the tables, each group at a learning rate proportional to its rms (Appendix~\ref{app:probe}; Tables~\ref{tab:trunk} and~\ref{tab:trunkdesign}). At a relative trunk step of $2.6\times10^{-4}$, about $10^{-5}$ in absolute terms for the matrices of the 1B model, the decomposition of the public run keeps its form: at 100k and 300k steps the standard lag is $+0.072$ and $+0.148$ nats, the step term $+0.033$ and $+0.068$, the reading term rises by $+0.163$ and $+0.209$ at context positions, and from resampled rows at matched step the old checkpoints are not behind (three rules met$^{\dagger}$). At a tenfold trunk step the form breaks: the standard lag is similar, but the step term almost disappears, the reading term no longer rises with age, and old checkpoints are behind even from resampled rows, by $+0.107\pm0.058$ and $+0.210\pm0.063$ nats (of four rules two failed and one was undecided$^{\dagger}$). The intervention does not produce this trunk-side lag: at the larger trunk step its pattern is intact (Table~\ref{tab:trunkdesign}), and from resampled rows the networks of \cG{} remain faster than those of \cN{} ($-0.057\pm0.008$). We therefore attribute it to the age of the public trunk rather than to the decay split; it is in line with the growing sensitivity of late checkpoints to weight perturbations \citep{springer2025overtrained} and lies outside our account.

\textbf{Decaying learning rate.} After 300k steps the learning rate of the public 1B run decays and the lag keeps growing, to $+0.375\pm0.003$ nats at the end. There $Z$ exceeds $Y$ by $+0.087\pm0.043$: part of the lag is no longer reproduced by the size of the table. Sharper minima at low learning rate are a documented source of fragility \citep{rofin2026learning,springer2025overtrained}. We restrict the account to the constant-rate phase.

\textbf{Runs, family, probe.} \LimRuns{} The intervention uses networks of width 1024 trained on $9\times10^7$ tokens, about 2{,}000 times fewer than the public 1B run has seen by 100k steps; their held-out loss is still falling, the lag inside them is measured over a stretch in which the learning rate falls, and we read differences between cells. Two features of the public runs are not reproduced by these networks: the arrangement part of the reading term (\S\ref{sec:public}), and a component shared by all rows on which the first layer comes to rely after the takeover (Appendix~\ref{app:mean}); whether the design causes them is open. The reading mismatch requires output normalization with decayed gains. Output normalization is used by OLMo~2 and the Gemma family, and whether gain vectors are decayed differs between implementations \citep{wang2026negligible,velikanov2026learnable}; the scale mismatch requires only an exempt table. Two models of other designs give no usable trend: they have one or two seeds, and in one of them the checkpoint-to-checkpoint fluctuations reach $0.25$ nats (Appendix~\ref{app:other}).

\section{Related work}\label{sec:related}

\textbf{Plasticity of language models.} \citet{springer2025overtrained} show that late checkpoints, of OLMo among others, are more sensitive to parameter perturbations and worse after fine-tuning, and note that in public checkpoints age is confounded with the learning rate; \citet{rofin2026learning} tie this sensitivity to learning-rate decay. Our main range has a constant learning rate, and the lag we find is a mismatch that two matchings remove; their fragility reappears in our data only when the trunk is moved fast (\S\ref{sec:limits}). \citet{han2026weight} show that larger weight decay in pretraining improves fine-tuning; they change the strength of decay, we change which groups are decayed. \citet{hernandezgarcia2026scale} study plasticity loss in language models trained on a stream of languages; \citet{chen2023improving} introduce embedding relearning as a measure of plasticity.

\textbf{Which parameters to decay.} That norm growth lowers the effective learning rate of scale-invariant weights is known \citep{vanlaarhoven2017l2}; \citet{lyle2024normalization} identify it as a source of plasticity loss, and \citet{liu2026effective} show that in language-model pretraining the loss follows the ratio of learning rate to parameter norm. Our scale mismatch is this mechanism met by a new embedding table; new are its source in a language model, the one group that is exempt from decay, shown by intervention, and its size. The balance $\sqrt{\eta/2\lambda}$ and the random-walk growth are those of \citet{kosson2024rotational} and \citet{kosson2026weight}, and \citet{velikanov2026learnable} stress that decay thereby fixes the scale of a weight by the optimizer rather than by the data. Which groups should be decayed is an open design question. Output normalization with learned gains is used by OLMo~2, by Gemma~2 \citep{gemma2024gemma2} and, as \citet{wang2026negligible} note, by Gemma~3; the same authors observe that implementations differ in whether they decay gain vectors, argue that decaying the gains of output normalization weakens a sublayer relative to the stream, and find a higher final loss with it in their networks. The designers of OLMo~2 exempted the table because a decayed table became too small for stable training \citep{olmo2024olmo2,takase2025spike}, and \citet{liu2026nearzero} recommend the exemption because decayed rows of unseen tokens collapse. None of these works follows what the exemption does next to decayed gains: the table and the gains move in opposite directions, their ratio decides what the trunk reads, and that decides how a new vocabulary can be learned.

\textbf{Initializing embeddings.} Starting new embeddings from the mean or from a Gaussian with the moments of the old table is established for vocabulary extension \citep{hewitt2021initializing,mundra2024empirical}, and a frozen trunk is known to expect embeddings in the subspace of the old ones \citep{downey2023embedding}. We add that the advantage of such a start is not constant: it opens with the share of the table in the stream and is negative in networks that have not switched.

\section{Conclusion}

In OLMo~2, relearning a new vocabulary slows with pretraining age because one parameter group, the embedding table, is exempt from weight decay while the gains around it are decayed. The table grows, so a fixed step shrinks relative to it; it outgrows the gains, so the trunk comes to read its rows as they are. A $2\times2$ intervention reproduces both mismatches and their growth with age; decaying the table closes the first and limits the second. With step and rows matched an older trunk relearns faster. A probe of plasticity that re-initializes parameters should scale its step to what it replaces and say which statistics its start preserves, or it measures the design, not the trunk; and which groups are decayed decides what the residual stream is made of.

\section*{AI Use Statement}
The authors formulated the research questions, hypotheses and claims. Generative AI assisted with experimental implementation, debugging, analysis, figures, reports, drafting, translation and editing. The authors verified the reported results, read and revised the manuscript, and take responsibility for the text, claims, code and experiments.

\bibliography{refs}
\bibliographystyle{plainnat}

\clearpage
\appendix
\section{Details of the probe and of the networks}\label{app:probe}

\textbf{Public checkpoints.} OLMo-2-0425-1B, stage 1, at 10k, 20k, 30k, 40k, 50k, 60k, 70k, 80k, 100k, 150k, 200k, 250k, 300k, 700k, 1.2M and 1{,}907{,}359 steps ($2{,}097{,}152$ tokens per step). The official configuration gives AdamW with $\beta=(0.9,0.95)$, weight decay $0.1$, \texttt{decay\_norm\_and\_bias: true}, \texttt{decay\_embeddings: false}, and a cosine schedule from $4\times10^{-4}$ over $5\times10^{12}$ tokens to $0.1$ of the peak after a linear warm-up of $8.4\times10^{9}$ tokens; stage 1 stops at $4\times10^{12}$ tokens. From this schedule the learning rate is $0.999$, $0.996$, $0.992$, $0.985$, $0.976$, $0.966$, $0.823$, $0.546$ and $0.186$ of the peak at 40k, 100k, 150k, 200k, 250k, 300k, 700k, 1.2M steps and at the end. \AppSeven

\textbf{Probe.} The text is English Wikipedia in sequences of 512 tokens. The vocabulary of a run is restricted to the token types that occur in its training and evaluation text (about $4\times10^4$ for the public models). The input table is replaced by a new table with one row per type and the output table by zeros; the gain of the final normalization is rescaled to rms 1. Both tables are trained for 300 updates of 8 sequences with Adam ($\beta=(0.9,0.95)$, $\epsilon=10^{-8}$, 15 warm-up updates, clipping of the global gradient norm at 1), weight decay $0.1$ on the output table and none on the input table. The held-out loss on 32 fixed sequences is recorded every 25 updates; the same 32 sequences are used for all seeds and checkpoints, so the standard errors over probe seeds do not contain the variation of the evaluation text. A position counts as a \emph{context} position if its target token occurred earlier in the same sequence and as a \emph{first-occurrence} position otherwise. The probe seed selects the training text and the random streams of the new table. For the public models the comparisons of the main text use probe seeds 2 to 5; for the own networks, seeds 2 to 5 at the standard relative step and seeds 2 and 3 on the step grids. Two runs of the same cell and seed differ by $0.004$ nats or less at the standard relative step (14 repeated pairs in the public 1B run).

\textbf{Populations for \rows{} and \gauss.} Old rows are drawn from 50{,}000 token types. For the own networks these are the most frequent types of the pretraining corpus. For the public models the types are ranked by their count in a sample of $5.1\times10^{5}$ tokens of the probe corpus; 31{,}733 types occur in the sample, and the remaining 18{,}267 are types with zero count taken in the order of the ranking routine. The mean and covariance of \gauss{} are those of the same population.

\textbf{Share.} For the public models $\tau$ is the ratio of the summed power of the raw rows to the summed power of the stream entering the second layer over the positions of 16 held-out sequences; for the own networks it is the mean over positions of the per-position ratio, as logged during pretraining. On the public 1B run the two conventions give $0.18$ and $0.18$ at 40k steps, $0.92$ and $0.88$ at 100k and $0.91$ and $0.85$ at 300k. Because the output of the first layer is not orthogonal to the row, the ratio can slightly exceed one.

\textbf{Own networks.} \texttt{Olmo2ForCausalLM} with 12 layers, width 1024, 16 heads, feed-forward width 4096, a 50{,}304-token BPE vocabulary trained on the corpus, sequences of 512 tokens, 8 sequences per update, AdamW with $\beta=(0.9,0.95)$, $\epsilon=10^{-8}$, weight decay $0.1$ on matrices, gradient clipping at 1, the auxiliary softmax loss of the official configuration ($10^{-5}$), normal initialization with standard deviation $0.02$, peak learning rate $2\times10^{-3}$ after 1{,}000 warm-up updates and a cosine to $0.1$ of the peak whose horizon is 50{,}000 updates ($1.25$ times a nominal length of 40{,}000, as the horizon of the official schedule exceeds its stage). Training is stopped at 22{,}000 updates; the held-out loss is still falling there (it is within $0.005$ of its minimum over the run in all networks). Before the runs we instantiated each cell on CPU and listed the parameters in each decay group: the cells differ from one another only in the weight decay of the input table ($0$ or $0.1$) and of the 24 output-gain vectors ($0$ or $0.1$).

\textbf{Training the trunk (\S\ref{sec:limits}).} All parameters of the trunk are trained together with the two tables. Each group (attention and feed-forward matrices, gate matrices, attention output gains, feed-forward output gains, query/key gains) receives the learning rate $s_{\mathrm{trunk}}$ times the rms of the group, and the final gain $s_{\mathrm{trunk}}$; matrices are decayed with $\lambda=0.1$ and gains are not; warm-up, clipping and the tables are as in the frozen probe. For the public model the weights are kept in bfloat16 with stochastic rounding, for the own networks in float32.

\section{Decay settings: the official flag and other networks}\label{app:arms}

\textbf{Why \cG{} and \cB{} decay only the output gains.} The official flag decays every normalization gain: the two output gains of each layer, the gains of the query and key normalization, and the gain of the final normalization. In a network of our size with the official flag, $49\%$ of the entries of the query/key gains collapse below $0.05$ by 22{,}000 updates, against $15\%$ at the end of the public 1B run and $21\%$ for the 7B model; in the public 1B run the collapsed fraction rises from $0.2\%$ at 40k steps to $7\%$ at 100k and $13\%$ at 300k and is concentrated in the first layer. To keep this second effect of the flag out of the comparison, \cG{} and \cB{} decay only the output gains (collapsed fraction $1.4$ to $2.0\%$). The complementary network, with decay only on the query/key and final gains, is not taken over ($\tau=0.050$ at 22{,}000 updates) and its moment part $D^{\mathrm{ctx}}$ is $-0.095$, as in \cN{} and \cT. The network with the official flag is taken over ($\tau=0.67$ at 16{,}000 and $0.76$ at 22{,}000 updates) and has $D^{\mathrm{ctx}}=+0.209$ and $+0.207$. A network with $\lambda=0.3$ on the output gains is taken over earlier ($\tau$ crosses $1/2$ near 6{,}300 updates instead of 11{,}400 to 12{,}400) and has $D^{\mathrm{ctx}}=+0.270$ at 10{,}000 and $+0.209$ at 22{,}000 updates. A network with the official flag at $\lambda=0.3$, trained to 10{,}000 updates, has $\tau=0.71$ and $D^{\mathrm{ctx}}=+0.262$ there, and a further run without gain decay has $\tau=0.011$ and $D^{\mathrm{ctx}}=-0.149$. These five networks have one pretraining seed each; they enter Figure~\ref{fig:design}c as ``other decay settings''. At equal $\tau$ near $0.35$, networks that stay there (\cB, $D^{\mathrm{ctx}}=+0.02$ to $+0.05$) are above networks passing through on the way up ($-0.06$ to $+0.01$): the trunk's adaptation trails the share.

\textbf{Gains in all layers.} At 22{,}000 updates the rms of the decayed output gains, in units of $\bal$, ranges over layers from $0.46$ to $2.04$ (attention) and $0.42$ to $3.23$ (feed-forward) in the three networks of \cG{} and from $0.50$ to $2.02$ and $0.41$ to $2.71$ in the three networks of \cB, with medians $1.01$ and $0.89$. The first attention gain is at $0.92$ to $1.04$; the first feed-forward gain is at $0.41$ to $0.46$; the gains of the last feed-forward blocks are the largest. Undecayed gains range from $4.3$ to $23$. The rows of a decayed table are at $0.96$ to $0.99$ times $\bal$ for frequent tokens, $0.69$ for tokens of intermediate frequency and $0.36$ for rare tokens; rows of tokens that never occur in the training text shrink to zero.

\textbf{Same trajectory of the gains with and without a decayed table.} The geometric mean of all output-gain entries is $0.487$ and $0.486$ at 3{,}000 updates, $0.064$ and $0.057$ at 10{,}000 and $0.0126$ and $0.0118$ at 22{,}000 in \cB{} and \cG{} (seed 0). The share differs because the table differs: at 10{,}000, 14{,}000, 18{,}000 and 22{,}000 updates $\tau$ is $0.128$, $0.239$, $0.287$, $0.299$ in \cB{} and $0.318$, $0.567$, $0.670$, $0.713$ in \cG.

\section{The random-walk coefficient}\label{app:walk}

We simulate one coordinate that receives a unit-variance, zero-mean gradient with probability $p$ at each update and a zero gradient otherwise, under Adam with $\beta=(0.9,0.95)$ and bias correction (4{,}000 coordinates, 20{,}000 updates, $\eta=10^{-3}$). Without weight decay the rms after $n$ updates is $c(p)\,\eta\sqrt n$ with $c=1.00$, $0.96$, $0.87$, $0.70$ and $0.51$ for $p=1$, $0.3$, $0.1$, $0.03$ and $0.01$. With $\lambda=0.1$ the rms settles at $0.98$, $0.95$, $0.86$, $0.70$ and $0.50$ times $\sqrt{\eta/2\lambda}$. The same coefficient thus governs the growth of an exempt row and the balance of a decayed one, and it is below one for rows of tokens that are absent from most batches. In the public 1B run $c$ of the whole table is $\NcA$, $\NcB$ and $\NcC$ at 40k, 100k and 300k steps (from the rms of the table, its initial value $0.0197$ and the official schedule); it is between $0.83$ and $0.85$ at every checkpoint up to 60k steps and falls afterwards, to $0.57$ at the end, so after the takeover the growth is slower than a free random walk.

\section{The component shared by all rows}\label{app:mean}

After the takeover the public 1B trunk comes to rely on a component that all rows of its table share, and the own networks do not reach this stage. We report the facts and leave the cause open.

\textbf{Relearning.} Let $M$ be the loss with resampled rows from which the mean row has been removed minus the loss with resampled rows, at matched relative step (four probe seeds, all positions). $M$ is $-0.007\pm0.010$ at 40k steps, $+0.013$, $+0.029$, $+0.035$ and $+0.032$ at 50k to 80k, and $+0.068\pm0.016$ at 100k; with eight probe seeds it is $-0.006$ at 40k, $+0.068$ at 100k and $+0.123$ at 300k steps.

\textbf{Forward pass.} Subtracting the unweighted mean of the population rows from every row raises the held-out loss of the unchanged model by $+0.02$ nats at 40k steps, $+0.09$ at 60k, $+0.39$ at 80k, $+0.72$ at 100k, $+1.17$ at 150k and $+1.75$ at 300k. In young checkpoints the frequency-weighted mean row is a direction that encodes token frequency rather than a component of all rows: the mean of the rows of tokens that never occur in our text has cosine $-0.24$ with it at 40k steps. After the takeover the rows of rare tokens align with it ($-0.01$ at 80k, $+0.08$ at 100k, $+0.25$ at 150k, $+0.49$ at 300k), and the first feed-forward block comes to pass this component on (the ratio of the shared to the token-specific part of its gate pre-activations rises from $0.26$ at 40k steps to $0.44$ at 100k and $0.64$ at 300k).

\textbf{Own networks.} At 22{,}000 updates subtracting the unweighted mean costs $+0.01$ nats in all cells, and the frequency-weighted mean matters with and without the takeover ($+0.08$ to $+0.25$ in \cG, $+0.23$ to $+0.25$ in \cN). These networks see about $2{,}000$ times fewer tokens than the public run has seen at 100k steps and do not reach the stage in which rare rows align; whether the decay of the gains causes that stage cannot be decided with them.

\section{Models with other designs}\label{app:other}

These measurements come from earlier stages of this project, with one or two probe seeds and partly with an earlier version of the probe; we list them for orientation and do not build on them.

\textbf{Pythia-1.4B} \citep{biderman2023pythia} is a pre-normalization model that decays its embedding table, whose rms does not grow ($0.023$, $0.026$, $0.026$ and $0.022$ at 16k, 33k, 66k and 143k steps). Under the standard probe its relearning loss shows no trend with age: $6.99$, $6.75$, $7.02$ and $6.77$ at the four checkpoints with one seed, and $+0.02$ and $+0.03$ nats for the 143k-step checkpoint against the 33k-step one with two seeds. The isotropic probe fluctuates by $0.25$ nats from checkpoint to checkpoint in this model. From resampled rows the 143k-step checkpoint is behind the 33k-step one ($+0.18$ and $+0.09$), a lag of another kind that this account does not cover.

\textbf{Amber} \citep{liu2023llm360}, a pre-normalization model, has a table whose rms does not keep growing ($0.028$, $0.034$, $0.036$ and $0.027$ at four checkpoints) and shows no lag with age when the embeddings are relearned through its lower 16 layers (one seed; the last checkpoint is $0.07$ nats ahead of the first). Both models are consistent with the scale mismatch requiring a growing table; neither is a test of the reading mismatch.

\section{Additional tables}\label{app:tables}

\begin{table}[h]
\caption{Public 1B run, frozen trunk: the terms of Eq.~\eqref{eq:ledger} over all positions and at context and first-occurrence positions (nats; mean $\pm$ standard error over four probe seeds). $\Delta D$: rise of the moment part; $R_{\textsc{moment}}$: as $R$ with moment-matched tables instead of resampled rows.}\label{tab:ledgerpos}
\centering\footnotesize
\fit{\begin{tabular}{@{}lllccccccc@{}}
\toprule
$\eta_{\mathrm{out}}$ & steps & pos. & $T$ & $Z$ & $Y$ & $\Delta S$ & $\Delta D$ & $R$ & $R_{\textsc{moment}}$ \\
\midrule
$10^{-3}$ & 100k & all & $+0.108{\pm}0.027$ & $+0.013{\pm}0.016$ & $+0.055{\pm}0.009$ & $+0.199{\pm}0.012$ & $+0.111{\pm}0.011$ & $-0.105{\pm}0.005$ & $-0.016{\pm}0.011$ \\
 &  & ctx & $+0.218{\pm}0.038$ & $-0.008{\pm}0.022$ & $+0.046{\pm}0.019$ & $+0.293{\pm}0.017$ & $+0.130{\pm}0.010$ & $-0.067{\pm}0.006$ & $+0.096{\pm}0.025$ \\
 &  & first & $+0.004{\pm}0.018$ & $+0.033{\pm}0.011$ & $+0.063{\pm}0.013$ & $+0.111{\pm}0.008$ & $+0.092{\pm}0.023$ & $-0.140{\pm}0.004$ & $-0.122{\pm}0.019$ \\
 & 150k & all & $+0.154{\pm}0.015$ & $+0.116{\pm}0.008$ & $+0.080{\pm}0.012$ & $+0.160{\pm}0.015$ & $+0.138{\pm}0.012$ & $-0.122{\pm}0.010$ & $-0.100{\pm}0.015$ \\
 &  & ctx & $+0.284{\pm}0.031$ & $+0.110{\pm}0.017$ & $+0.078{\pm}0.024$ & $+0.278{\pm}0.022$ & $+0.235{\pm}0.026$ & $-0.105{\pm}0.013$ & $-0.062{\pm}0.027$ \\
 &  & first & $+0.032{\pm}0.009$ & $+0.122{\pm}0.008$ & $+0.083{\pm}0.013$ & $+0.048{\pm}0.008$ & $+0.046{\pm}0.012$ & $-0.138{\pm}0.012$ & $-0.137{\pm}0.016$ \\
 & 200k & all & $+0.198{\pm}0.012$ & $+0.136{\pm}0.024$ & $+0.098{\pm}0.011$ & $+0.149{\pm}0.031$ & $+0.094{\pm}0.025$ & $-0.087{\pm}0.012$ & $-0.032{\pm}0.031$ \\
 &  & ctx & $+0.322{\pm}0.018$ & $+0.131{\pm}0.035$ & $+0.091{\pm}0.021$ & $+0.234{\pm}0.044$ & $+0.135{\pm}0.028$ & $-0.042{\pm}0.022$ & $+0.056{\pm}0.032$ \\
 &  & first & $+0.080{\pm}0.013$ & $+0.140{\pm}0.022$ & $+0.105{\pm}0.018$ & $+0.069{\pm}0.024$ & $+0.055{\pm}0.026$ & $-0.129{\pm}0.004$ & $-0.115{\pm}0.030$ \\
 & 250k & all & $+0.213{\pm}0.020$ & $+0.135{\pm}0.014$ & $+0.109{\pm}0.011$ & $+0.160{\pm}0.021$ & $+0.118{\pm}0.023$ & $-0.082{\pm}0.012$ & $-0.041{\pm}0.017$ \\
 &  & ctx & $+0.387{\pm}0.029$ & $+0.118{\pm}0.015$ & $+0.097{\pm}0.020$ & $+0.274{\pm}0.040$ & $+0.186{\pm}0.031$ & $-0.005{\pm}0.022$ & $+0.083{\pm}0.020$ \\
 &  & first & $+0.048{\pm}0.028$ & $+0.151{\pm}0.024$ & $+0.119{\pm}0.020$ & $+0.052{\pm}0.013$ & $+0.054{\pm}0.026$ & $-0.155{\pm}0.003$ & $-0.157{\pm}0.020$ \\
 & 300k & all & $+0.221{\pm}0.017$ & $+0.112{\pm}0.007$ & $+0.119{\pm}0.010$ & $+0.220{\pm}0.010$ & $+0.105{\pm}0.020$ & $-0.111{\pm}0.010$ & $+0.004{\pm}0.026$ \\
 &  & ctx & $+0.371{\pm}0.033$ & $+0.065{\pm}0.010$ & $+0.107{\pm}0.021$ & $+0.394{\pm}0.021$ & $+0.148{\pm}0.050$ & $-0.088{\pm}0.012$ & $+0.158{\pm}0.053$ \\
 &  & first & $+0.079{\pm}0.008$ & $+0.157{\pm}0.022$ & $+0.130{\pm}0.019$ & $+0.055{\pm}0.010$ & $+0.064{\pm}0.021$ & $-0.133{\pm}0.010$ & $-0.142{\pm}0.015$ \\
$3.3\!\times\!10^{-4}$ & 100k & all & $+0.024{\pm}0.015$ & $+0.055{\pm}0.007$ & $+0.058{\pm}0.004$ & $+0.095{\pm}0.012$ & $+0.075{\pm}0.010$ & $-0.125{\pm}0.008$ & $-0.106{\pm}0.009$ \\
 &  & ctx & $+0.033{\pm}0.021$ & $+0.059{\pm}0.019$ & $+0.058{\pm}0.004$ & $+0.090{\pm}0.024$ & $+0.057{\pm}0.034$ & $-0.116{\pm}0.014$ & $-0.083{\pm}0.010$ \\
 &  & first & $+0.015{\pm}0.010$ & $+0.050{\pm}0.013$ & $+0.058{\pm}0.008$ & $+0.099{\pm}0.008$ & $+0.092{\pm}0.016$ & $-0.134{\pm}0.008$ & $-0.127{\pm}0.024$ \\
 & 300k & all & $+0.071{\pm}0.015$ & $+0.154{\pm}0.010$ & $+0.127{\pm}0.006$ & $+0.093{\pm}0.016$ & $+0.074{\pm}0.015$ & $-0.176{\pm}0.004$ & $-0.157{\pm}0.015$ \\
 &  & ctx & $+0.118{\pm}0.008$ & $+0.141{\pm}0.011$ & $+0.118{\pm}0.006$ & $+0.141{\pm}0.028$ & $+0.082{\pm}0.015$ & $-0.164{\pm}0.016$ & $-0.106{\pm}0.017$ \\
 &  & first & $+0.026{\pm}0.024$ & $+0.165{\pm}0.009$ & $+0.135{\pm}0.013$ & $+0.048{\pm}0.029$ & $+0.067{\pm}0.023$ & $-0.187{\pm}0.019$ & $-0.206{\pm}0.020$ \\
\bottomrule
\end{tabular}
}
\end{table}

\begin{table}[h]
\caption{The intervention at two output learning rates of the probe (22{,}000 updates; relative step $0.0147$; frozen trunk; nats; mean $\pm$ standard error over three pretraining seeds, four probe seeds each). Upper block: moment part $D^{\mathrm{ctx}}$ per cell. Middle block: effect of one choice on $D^{\mathrm{ctx}}$ at fixed other choice.}\label{tab:designout}
\centering\small
\fit{\begin{tabular}{@{}lcc@{}}
\toprule
 & $\eta_{\mathrm{out}}=10^{-3}$ & $\eta_{\mathrm{out}}=3.3\times10^{-4}$ \\
\midrule
\cN & $-0.076\pm0.006$ & $-0.070\pm0.010$ \\
\cG & $+0.207\pm0.016$ & $+0.041\pm0.009$ \\
\cT & $-0.141\pm0.009$ & $-0.129\pm0.012$ \\
\cB & $+0.035\pm0.010$ & $-0.022\pm0.014$ \\
\midrule
decaying the gains, table exempt (\cG$-$\cN) & $+0.283\pm0.010$ & $+0.111\pm0.001$ \\
decaying the gains, table decayed (\cB$-$\cT) & $+0.176\pm0.018$ & $+0.106\pm0.022$ \\
decaying the table, gains not (\cT$-$\cN) & $-0.065\pm0.015$ & $-0.059\pm0.006$ \\
decaying the table, gains decayed (\cB$-$\cG) & $-0.172\pm0.012$ & $-0.063\pm0.017$ \\
\midrule
reading term $S^{\mathrm{ctx}}$: \cG$-$\cN & $+0.252\pm0.010$ & $+0.105\pm0.020$ \\
reading term $S^{\mathrm{ctx}}$: \cB$-$\cT & $+0.160\pm0.005$ & $+0.103\pm0.012$ \\
\rows, all positions: \cG$-$\cN & $-0.138\pm0.010$ & $-0.113\pm0.010$ \\
\iso, context: \cG$-$\cN & $+0.147\pm0.013$ & $+0.010\pm0.044$ \\
\bottomrule
\end{tabular}
}
\end{table}

\begin{table}[h]
\caption{Step grids of the twelve own networks at 22{,}000 updates (\iso; context positions; mean of probe seeds 2 and 3; output learning rate $10^{-3}$; frozen trunk). $\sigma_e$: rms of the table. $\ell=\sigma_e/\sqrt\tau$: rms of the stream entering the second layer. $s^\ast$: relative step with the lowest loss on the grid $\{0.0147, 0.045, 0.08, 0.13, 0.2, 0.4, 1.2\}$ (and $2.4$ for \cT{} and \cB; $^\ast$: at the upper end of the grid); $\eta^\ast=s^\ast\sigma_e$; $L^{\mathrm{ctx}\ast}$: loss at $s^\ast$.}\label{tab:best}
\centering\small
\fit{\begin{tabular}{@{}lccccccc@{}}
\toprule
cell & seed & $\sigma_e$ & $\ell$ & $s^\ast$ & $\eta^\ast$ & $\eta^\ast/\ell$ & $L^{\mathrm{ctx}\ast}$ \\
\midrule
\cN & 0 & $0.1469$ & $0.679$ & $0.4$ & $0.0588$ & $0.09$ & $4.538$ \\
 & 1 & $0.1469$ & $0.664$ & $0.4$ & $0.0588$ & $0.09$ & $4.612$ \\
 & 2 & $0.1468$ & $0.644$ & $0.4$ & $0.0587$ & $0.09$ & $4.576$ \\
\cG & 0 & $0.1494$ & $0.177$ & $0.08$ & $0.0120$ & $0.07$ & $4.610$ \\
 & 1 & $0.1496$ & $0.168$ & $0.08$ & $0.0120$ & $0.07$ & $4.689$ \\
 & 2 & $0.1495$ & $0.167$ & $0.13$ & $0.0194$ & $0.12$ & $4.660$ \\
\cT & 0 & $0.0478$ & $0.564$ & $2.4$ & $0.1147$ & $0.20$$^\ast$ & $4.556$ \\
 & 1 & $0.0478$ & $0.571$ & $1.2$ & $0.0574$ & $0.10$ & $4.563$ \\
 & 2 & $0.0478$ & $0.535$ & $1.2$ & $0.0573$ & $0.11$ & $4.615$ \\
\cB & 0 & $0.0500$ & $0.091$ & $0.2$ & $0.0100$ & $0.11$ & $4.556$ \\
 & 1 & $0.0500$ & $0.087$ & $0.2$ & $0.0100$ & $0.11$ & $4.564$ \\
 & 2 & $0.0500$ & $0.089$ & $0.2$ & $0.0100$ & $0.11$ & $4.538$ \\
\bottomrule
\end{tabular}
}
\end{table}

\begin{table}[h]
\caption{The lag inside each network of the intervention, by position type: the terms of Eq.~\eqref{eq:ledger} between the checkpoints at 10{,}000 and 22{,}000 updates of the same network (frozen trunk; output learning rate $10^{-3}$; nats; mean $\pm$ standard error over three pretraining seeds, four probe seeds each). The old side at the absolute learning rate of the young side gives $T$ and $Z$; both sides at relative step $0.0147$ give $S$, $D$ and $R$. A repetition of 48 runs of an earlier stage at 10{,}000 updates differs by at most $0.006$ nats. Measured against a 3{,}000-update network instead (held-out loss $5.2$), the loss at first-occurrence positions rises with age by $+0.088$ and $+0.090$ nats in \cN{} and \cG{} and by $-0.014$ and $-0.015$ in \cT{} and \cB.}
\label{tab:within}
\centering\small
\fit{\begin{tabular}{@{}llccccc@{}}
\toprule
pos. & cell & $T$ & $Z$ & $\Delta S$ & $\Delta D$ & $R$ \\
\midrule
all & \cN & $+0.053{\pm}0.024$ & $+0.037{\pm}0.002$ & $+0.032{\pm}0.010$ & $+0.035{\pm}0.011$ & $-0.016{\pm}0.014$ \\
 & \cG & $+0.097{\pm}0.003$ & $+0.084{\pm}0.028$ & $+0.119{\pm}0.024$ & $+0.143{\pm}0.014$ & $-0.106{\pm}0.002$ \\
 & \cT & $+0.023{\pm}0.016$ & $-0.011{\pm}0.003$ & $+0.030{\pm}0.010$ & $+0.030{\pm}0.015$ & $+0.003{\pm}0.009$ \\
 & \cB & $+0.036{\pm}0.016$ & $-0.014{\pm}0.002$ & $+0.100{\pm}0.016$ & $+0.094{\pm}0.014$ & $-0.051{\pm}0.004$ \\
\midrule
ctx & \cN & $+0.054{\pm}0.042$ & $+0.031{\pm}0.004$ & $+0.043{\pm}0.015$ & $+0.039{\pm}0.012$ & $-0.020{\pm}0.030$ \\
 & \cG & $+0.209{\pm}0.012$ & $+0.079{\pm}0.035$ & $+0.211{\pm}0.037$ & $+0.247{\pm}0.020$ & $-0.081{\pm}0.002$ \\
 & \cT & $+0.018{\pm}0.027$ & $-0.009{\pm}0.006$ & $+0.039{\pm}0.012$ & $+0.030{\pm}0.038$ & $-0.011{\pm}0.021$ \\
 & \cB & $+0.084{\pm}0.034$ & $-0.012{\pm}0.003$ & $+0.147{\pm}0.019$ & $+0.127{\pm}0.010$ & $-0.052{\pm}0.016$ \\
\midrule
first & \cN & $+0.051{\pm}0.007$ & $+0.043{\pm}0.006$ & $+0.021{\pm}0.007$ & $+0.030{\pm}0.010$ & $-0.013{\pm}0.003$ \\
 & \cG & $-0.014{\pm}0.013$ & $+0.089{\pm}0.021$ & $+0.027{\pm}0.021$ & $+0.041{\pm}0.015$ & $-0.130{\pm}0.004$ \\
 & \cT & $+0.028{\pm}0.007$ & $-0.013{\pm}0.002$ & $+0.022{\pm}0.011$ & $+0.029{\pm}0.009$ & $+0.018{\pm}0.014$ \\
 & \cB & $-0.012{\pm}0.010$ & $-0.016{\pm}0.001$ & $+0.053{\pm}0.018$ & $+0.062{\pm}0.017$ & $-0.050{\pm}0.012$ \\
\bottomrule
\end{tabular}
}
\end{table}

\begin{table}[h]
\caption{Public 1B run, frozen trunk, output learning rate $10^{-3}$: the two terms of Eq.~\eqref{eq:ledger} in both orders of matching (nats; mean $\pm$ standard error over four probe seeds). Step first: $Z$ from isotropic tables, $\Delta S$ at matched step. Table first: $\Delta S'$ at the standard step, $Z'$ from resampled rows. $T$ and $R$ are the same in both orders, and $Z'-Z=\Delta S-\Delta S'$.}
\label{tab:order}
\centering\small
\fit{\begin{tabular}{@{}llccccc@{}}
\toprule
 & & \multicolumn{2}{c}{step first (Eq.~\ref{eq:ledger})} & \multicolumn{2}{c}{table first} & \\
steps & pos. & $Z$ & $\Delta S$ & $Z'$ & $\Delta S'$ & $Z'-Z$ \\
\midrule
100k & all & $+0.013{\pm}0.016$ & $+0.199{\pm}0.012$ & $+0.082{\pm}0.013$ & $+0.131{\pm}0.022$ & $+0.068{\pm}0.010$ \\
 & ctx & $-0.008{\pm}0.022$ & $+0.293{\pm}0.017$ & $+0.068{\pm}0.016$ & $+0.218{\pm}0.028$ & $+0.075{\pm}0.011$ \\
 & first & $+0.033{\pm}0.011$ & $+0.111{\pm}0.008$ & $+0.095{\pm}0.011$ & $+0.049{\pm}0.017$ & $+0.061{\pm}0.009$ \\
\addlinespace[2pt]
150k & all & $+0.116{\pm}0.008$ & $+0.160{\pm}0.015$ & $+0.115{\pm}0.011$ & $+0.161{\pm}0.017$ & $-0.001{\pm}0.007$ \\
 & ctx & $+0.110{\pm}0.017$ & $+0.278{\pm}0.022$ & $+0.083{\pm}0.008$ & $+0.305{\pm}0.030$ & $-0.027{\pm}0.009$ \\
 & first & $+0.122{\pm}0.008$ & $+0.048{\pm}0.008$ & $+0.145{\pm}0.019$ & $+0.025{\pm}0.014$ & $+0.023{\pm}0.014$ \\
\addlinespace[2pt]
200k & all & $+0.136{\pm}0.024$ & $+0.149{\pm}0.031$ & $+0.128{\pm}0.013$ & $+0.156{\pm}0.006$ & $-0.007{\pm}0.031$ \\
 & ctx & $+0.131{\pm}0.035$ & $+0.234{\pm}0.044$ & $+0.071{\pm}0.031$ & $+0.293{\pm}0.018$ & $-0.060{\pm}0.055$ \\
 & first & $+0.140{\pm}0.022$ & $+0.069{\pm}0.024$ & $+0.182{\pm}0.006$ & $+0.027{\pm}0.014$ & $+0.042{\pm}0.019$ \\
\addlinespace[2pt]
250k & all & $+0.135{\pm}0.014$ & $+0.160{\pm}0.021$ & $+0.122{\pm}0.015$ & $+0.173{\pm}0.017$ & $-0.013{\pm}0.008$ \\
 & ctx & $+0.118{\pm}0.015$ & $+0.274{\pm}0.040$ & $+0.056{\pm}0.022$ & $+0.336{\pm}0.022$ & $-0.062{\pm}0.024$ \\
 & first & $+0.151{\pm}0.024$ & $+0.052{\pm}0.013$ & $+0.185{\pm}0.010$ & $+0.019{\pm}0.021$ & $+0.033{\pm}0.014$ \\
\addlinespace[2pt]
300k & all & $+0.112{\pm}0.007$ & $+0.220{\pm}0.010$ & $+0.163{\pm}0.008$ & $+0.168{\pm}0.016$ & $+0.051{\pm}0.007$ \\
 & ctx & $+0.065{\pm}0.010$ & $+0.394{\pm}0.021$ & $+0.137{\pm}0.007$ & $+0.322{\pm}0.022$ & $+0.072{\pm}0.008$ \\
 & first & $+0.157{\pm}0.022$ & $+0.055{\pm}0.010$ & $+0.189{\pm}0.020$ & $+0.023{\pm}0.011$ & $+0.032{\pm}0.017$ \\
\bottomrule
\end{tabular}
}
\end{table}

\begin{table}[h]
\caption{Public 1B run across the takeover (context positions; nats; four probe seeds; matched relative step). $S=D+H$ as in Table~\ref{tab:glossary}.}
\label{tab:switch}
\centering\small
\fit{\begin{tabular}{@{}lcccc@{}}
\toprule
steps & $\tau$ & $S^{\mathrm{ctx}}$ & $D^{\mathrm{ctx}}$ & $H^{\mathrm{ctx}}$ \\
\midrule
40k & $0.18$ & $+0.149\pm0.011$ & $+0.105$ & $+0.044$ \\
50k & $0.47$ & $+0.243\pm0.013$ & $+0.159$ & $+0.083$ \\
60k & $0.68$ & $+0.212\pm0.010$ & $+0.061$ & $+0.151$ \\
70k & $0.80$ & $+0.269\pm0.035$ & $+0.122$ & $+0.147$ \\
80k & $0.86$ & $+0.334\pm0.042$ & $+0.157$ & $+0.177$ \\
100k & $0.92$ & $+0.443\pm0.012$ & $+0.235$ & $+0.207$ \\
\bottomrule
\end{tabular}}
\end{table}

\begin{table}[h]
\caption{Public 1B run, frozen trunk: relearning loss on grids of the relative step $s$ (nats; mean of probe seeds 2 and 3). Bold: lowest loss of the row. The rules were fixed on the grid from $s=0.0147$ to $0.2$ ($0.4$ at 40k steps), over all positions; $s=0.8$ at 40k steps, $s=0.4$ at 100k steps and, for resampled rows, $s=0.005$ and $0.0085$ were added afterwards so that the optima over all positions and at context positions lie inside the grid (at first-occurrence positions the optimum of the isotropic table at 40k steps is at the upper edge). The optima over all positions did not change. Differences quoted in the text are computed before rounding.}\label{tab:grid}
\centering\small
\fit{\begin{tabular}{@{}lllccccccccc@{}}
\toprule
table & steps & pos. & $s=0.005$ & $0.0085$ & $0.0147$ & $0.045$ & $0.08$ & $0.13$ & $0.2$ & $0.4$ & $0.8$ \\
\midrule
\rows & 40k & all & $6.890$ & $6.773$ & $6.665$ & $6.488$ & $6.505$ & $6.481$ & $6.478$ & $\mathbf{6.463}$ & $6.556$ \\
 &  & ctx & $4.700$ & $4.604$ & $4.526$ & $\mathbf{4.399}$ & $4.460$ & $4.466$ & $4.491$ & $4.534$ & $4.694$ \\
 &  & first & $8.956$ & $8.818$ & $8.681$ & $8.457$ & $8.434$ & $8.381$ & $8.353$ & $\mathbf{8.282}$ & $8.313$ \\
 & 100k & all & $6.809$ & $6.677$ & $6.553$ & $6.491$ & $6.484$ & $\mathbf{6.467}$ & $6.516$ & $6.719$ &  \\
 &  & ctx & $4.659$ & $4.542$ & $\mathbf{4.450}$ & $4.481$ & $4.525$ & $4.545$ & $4.653$ & $4.961$ &  \\
 &  & first & $8.837$ & $8.690$ & $8.535$ & $8.387$ & $8.331$ & $8.279$ & $\mathbf{8.273}$ & $8.377$ &  \\
 & 300k & all & $6.780$ & $6.647$ & $6.542$ & $\mathbf{6.471}$ & $6.493$ & $6.523$ & $6.652$ &  &  \\
 &  & ctx & $4.619$ & $4.499$ & $\mathbf{4.429}$ & $4.487$ & $4.571$ & $4.627$ & $4.843$ &  &  \\
 &  & first & $8.817$ & $8.672$ & $8.535$ & $8.341$ & $\mathbf{8.305}$ & $8.311$ & $8.359$ &  &  \\
\midrule
\iso & 40k & all &  &  & $6.709$ & $6.650$ & $6.607$ & $6.591$ & $6.561$ & $\mathbf{6.528}$ & $6.576$ \\
 &  & ctx &  &  & $4.688$ & $4.685$ & $4.661$ & $4.653$ & $4.641$ & $\mathbf{4.633}$ & $4.767$ \\
 &  & first &  &  & $8.615$ & $8.504$ & $8.442$ & $8.418$ & $8.371$ & $8.315$ & $\mathbf{8.281}$ \\
 & 100k & all &  &  & $6.782$ & $6.630$ & $6.611$ & $6.585$ & $\mathbf{6.543}$ & $6.669$ &  \\
 &  & ctx &  &  & $4.887$ & $4.751$ & $4.782$ & $4.767$ & $\mathbf{4.745}$ & $4.926$ &  \\
 &  & first &  &  & $8.570$ & $8.401$ & $8.336$ & $8.299$ & $\mathbf{8.239}$ & $8.313$ &  \\
 & 300k & all &  &  & $6.807$ & $6.662$ & $\mathbf{6.647}$ & $6.654$ & $6.682$ &  &  \\
 &  & ctx &  &  & $4.987$ & $4.835$ & $\mathbf{4.824}$ & $4.868$ & $4.918$ &  &  \\
 &  & first &  &  & $8.523$ & $8.385$ & $8.367$ & $\mathbf{8.337}$ & $8.346$ &  &  \\
\bottomrule
\end{tabular}
}
\end{table}

\begin{table}[h]
\caption{Public 1B run with the trunk frozen and with the trunk trained at two relative steps (nats; mean $\pm$ standard error over four probe seeds). Columns as in Table~\ref{tab:ledger}.}\label{tab:trunk}
\centering\footnotesize
\fit{\begin{tabular}{@{}lllccccc@{}}
\toprule
pos. & trunk step & steps & $T$ & $Z$ & $Y$ & $\Delta S$ & $R$ \\
\midrule
all & frozen & 100k & $+0.108{\pm}0.027$ & $+0.013{\pm}0.016$ & $+0.055{\pm}0.009$ & $+0.199{\pm}0.012$ & $-0.105{\pm}0.005$ \\
 &  & 300k & $+0.221{\pm}0.017$ & $+0.112{\pm}0.007$ & $+0.119{\pm}0.010$ & $+0.220{\pm}0.010$ & $-0.111{\pm}0.010$ \\
 & $2.6\times10^{-4}$ & 100k & $+0.072{\pm}0.029$ & $+0.033{\pm}0.016$ & $+0.026{\pm}0.020$ & $+0.043{\pm}0.031$ & $-0.004{\pm}0.006$ \\
 &  & 300k & $+0.148{\pm}0.016$ & $+0.068{\pm}0.017$ & $+0.050{\pm}0.023$ & $+0.104{\pm}0.014$ & $-0.024{\pm}0.008$ \\
 & $2.6\times10^{-3}$ & 100k & $+0.068{\pm}0.021$ & $+0.010{\pm}0.004$ & $-0.018{\pm}0.016$ & $-0.049{\pm}0.046$ & $+0.107{\pm}0.058$ \\
 &  & 300k & $+0.166{\pm}0.014$ & $+0.024{\pm}0.010$ & $+0.024{\pm}0.018$ & $-0.067{\pm}0.075$ & $+0.210{\pm}0.063$ \\
\midrule
ctx & frozen & 100k & $+0.218{\pm}0.038$ & $-0.008{\pm}0.022$ & $+0.046{\pm}0.019$ & $+0.293{\pm}0.017$ & $-0.067{\pm}0.006$ \\
 &  & 300k & $+0.371{\pm}0.033$ & $+0.065{\pm}0.010$ & $+0.107{\pm}0.021$ & $+0.394{\pm}0.021$ & $-0.088{\pm}0.012$ \\
 & $2.6\times10^{-4}$ & 100k & $+0.227{\pm}0.050$ & $+0.044{\pm}0.023$ & $+0.037{\pm}0.022$ & $+0.163{\pm}0.059$ & $+0.021{\pm}0.007$ \\
 &  & 300k & $+0.272{\pm}0.037$ & $+0.063{\pm}0.029$ & $+0.063{\pm}0.035$ & $+0.209{\pm}0.030$ & $-0.000{\pm}0.005$ \\
 & $2.6\times10^{-3}$ & 100k & $+0.146{\pm}0.031$ & $+0.026{\pm}0.009$ & $-0.028{\pm}0.026$ & $-0.097{\pm}0.106$ & $+0.218{\pm}0.123$ \\
 &  & 300k & $+0.281{\pm}0.012$ & $+0.031{\pm}0.015$ & $+0.033{\pm}0.035$ & $-0.130{\pm}0.155$ & $+0.381{\pm}0.133$ \\
\midrule
first & frozen & 100k & $+0.004{\pm}0.018$ & $+0.033{\pm}0.011$ & $+0.063{\pm}0.013$ & $+0.111{\pm}0.008$ & $-0.140{\pm}0.004$ \\
 &  & 300k & $+0.079{\pm}0.008$ & $+0.157{\pm}0.022$ & $+0.130{\pm}0.019$ & $+0.055{\pm}0.010$ & $-0.133{\pm}0.010$ \\
 & $2.6\times10^{-4}$ & 100k & $-0.074{\pm}0.018$ & $+0.023{\pm}0.012$ & $+0.015{\pm}0.020$ & $-0.069{\pm}0.015$ & $-0.028{\pm}0.007$ \\
 &  & 300k & $+0.031{\pm}0.006$ & $+0.073{\pm}0.013$ & $+0.037{\pm}0.012$ & $+0.005{\pm}0.006$ & $-0.047{\pm}0.020$ \\
 & $2.6\times10^{-3}$ & 100k & $-0.006{\pm}0.013$ & $-0.004{\pm}0.004$ & $-0.009{\pm}0.009$ & $-0.004{\pm}0.013$ & $+0.002{\pm}0.008$ \\
 &  & 300k & $+0.058{\pm}0.019$ & $+0.017{\pm}0.007$ & $+0.016{\pm}0.010$ & $-0.008{\pm}0.017$ & $+0.048{\pm}0.010$ \\
\bottomrule
\end{tabular}
}
\end{table}

\begin{table}[h]
\caption{The intervention with the trunk frozen and with the trunk trained at relative step $2.6\times10^{-3}$ (22{,}000 updates; nats; mean $\pm$ standard error over three pretraining seeds). Upper block: cells; lower block: effect of one choice at fixed other choice.}\label{tab:trunkdesign}
\centering\footnotesize
\fit{\begin{tabular}{@{}lcccccc@{}}
\toprule
 & \multicolumn{2}{c}{step term} & \multicolumn{2}{c}{$S^{\mathrm{ctx}}$} & \multicolumn{2}{c}{$D^{\mathrm{ctx}}$} \\
 & frozen & trained & frozen & trained & frozen & trained \\
\midrule
\cN & $+0.137{\pm}0.004$ & $+0.117{\pm}0.007$ & $-0.070{\pm}0.025$ & $-0.050{\pm}0.024$ & $-0.076{\pm}0.006$ & $-0.063{\pm}0.022$ \\
\cG & $+0.228{\pm}0.055$ & $+0.127{\pm}0.001$ & $+0.181{\pm}0.021$ & $+0.133{\pm}0.019$ & $+0.207{\pm}0.016$ & $+0.125{\pm}0.024$ \\
\cT & $+0.007{\pm}0.002$ & $+0.007{\pm}0.001$ & $-0.112{\pm}0.010$ & $-0.110{\pm}0.016$ & $-0.141{\pm}0.009$ & $-0.147{\pm}0.006$ \\
\cB & $+0.017{\pm}0.002$ & $+0.011{\pm}0.001$ & $+0.048{\pm}0.014$ & $+0.031{\pm}0.013$ & $+0.035{\pm}0.010$ & $+0.003{\pm}0.011$ \\
\midrule
\cT$-$\cN & $-0.130{\pm}0.004$ & $-0.110{\pm}0.007$ & $-0.042{\pm}0.015$ & $-0.060{\pm}0.014$ & $-0.065{\pm}0.015$ & $-0.084{\pm}0.016$ \\
\cB$-$\cG & $-0.211{\pm}0.056$ & $-0.116{\pm}0.002$ & $-0.134{\pm}0.011$ & $-0.103{\pm}0.009$ & $-0.172{\pm}0.012$ & $-0.123{\pm}0.014$ \\
\cG$-$\cN & $+0.090{\pm}0.058$ & $+0.010{\pm}0.007$ & $+0.252{\pm}0.010$ & $+0.183{\pm}0.019$ & $+0.283{\pm}0.010$ & $+0.189{\pm}0.032$ \\
\cB$-$\cT & $+0.010{\pm}0.004$ & $+0.004{\pm}0.001$ & $+0.160{\pm}0.005$ & $+0.141{\pm}0.011$ & $+0.176{\pm}0.018$ & $+0.150{\pm}0.012$ \\
\bottomrule
\end{tabular}
}
\end{table}

\begin{table}[h]
\caption{Public 7B run, frozen trunk, output learning rate $3.3\times10^{-4}$: the terms of Eq.~\eqref{eq:ledger} relative to the 30k-step checkpoint (nats; mean $\pm$ standard error over four probe seeds). Columns as in Table~\ref{tab:ledgerpos}.}\label{tab:seven}
\centering\footnotesize
\fit{\begin{tabular}{@{}llccccccc@{}}
\toprule
steps & pos. & $T$ & $Z$ & $Y$ & $\Delta S$ & $\Delta D$ & $R$ & $R_{\textsc{moment}}$ \\
\midrule
110k & all & $+0.096{\pm}0.046$ & $+0.135{\pm}0.032$ & $+0.053{\pm}0.040$ & $+0.152{\pm}0.034$ & $+0.069{\pm}0.034$ & $-0.190{\pm}0.017$ & $-0.107{\pm}0.008$ \\
 & ctx & $+0.265{\pm}0.067$ & $+0.142{\pm}0.042$ & $+0.054{\pm}0.054$ & $+0.226{\pm}0.054$ & $+0.049{\pm}0.042$ & $-0.103{\pm}0.015$ & $+0.074{\pm}0.014$ \\
 & first & $-0.062{\pm}0.030$ & $+0.128{\pm}0.027$ & $+0.052{\pm}0.028$ & $+0.082{\pm}0.020$ & $+0.087{\pm}0.026$ & $-0.273{\pm}0.021$ & $-0.278{\pm}0.011$ \\
200k & all & $+0.087{\pm}0.040$ & $+0.102{\pm}0.022$ & $+0.122{\pm}0.047$ & $+0.127{\pm}0.039$ & $+0.068{\pm}0.048$ & $-0.142{\pm}0.030$ & $-0.083{\pm}0.036$ \\
 & ctx & $+0.261{\pm}0.053$ & $+0.066{\pm}0.028$ & $+0.128{\pm}0.060$ & $+0.162{\pm}0.061$ & $+0.049{\pm}0.076$ & $+0.033{\pm}0.047$ & $+0.146{\pm}0.052$ \\
 & first & $-0.077{\pm}0.033$ & $+0.136{\pm}0.022$ & $+0.116{\pm}0.037$ & $+0.093{\pm}0.027$ & $+0.087{\pm}0.028$ & $-0.306{\pm}0.023$ & $-0.300{\pm}0.030$ \\
\bottomrule
\end{tabular}
}
\end{table}

\section{Rules fixed before the runs, and their outcomes}\label{app:prereg}

Every batch of runs had a written rule for how its result would be read, with thresholds, fixed and hashed before launch. Of the \NruleAll{} rules behind the results of this paper, \NruleOK{} came out as expected, \NruleFail{} failed and \NruleUndec{} were undecided by their own thresholds. We list the ones that did not come out as expected. \emph{Failed} means that the result fell on the side that the rule had fixed as rejecting; \emph{undecided} means that it fell between the two thresholds of the rule.
\begin{enumerate}
\item \emph{Failed.} ``With the table decayed and the gains decayed, the gains keep shrinking until $\tau$ exceeds $1/2$.'' Observed: $\tau=0.30$ at 22{,}000 updates and flat (seed 0). After this failure we fixed, for the two remaining seeds, the prediction that the first attention gain and the frequent rows stop within $0.85$ to $1.15$ times $\bal$ and that $\tau$ stops between $0.20$ and $0.40$; all four checks were met.
\item \emph{Failed.} ``If $\tau\le0.3$ in \cB, the moment part does not open ($<+0.10$ above \cT).'' Observed: $+0.148$ (seed 0). The binary reading (taken over or not) was too coarse; the graded reading of Figure~\ref{fig:design}c was then fixed for the other seeds and for six further checkpoints and was met.
\item \emph{Failed.} ``At the best step of each network, the difference of the moment part between \cG{} and \cN{} is at least $+0.03$ in all seeds.'' Observed: $+0.019$, $+0.078$, $+0.073$.
\item \emph{Failed.} ``At a third of the output learning rate, the effect of decaying the gains on the moment part with the table decayed is $0.3$ to $0.9$ of the effect with the table exempt.'' Observed: ratio $0.96$ ($+0.106$ against $+0.111$).
\item \emph{Undecided.} ``At the best step of each network, an isotropic table is at least $0.10$ nats worse at context positions in \cG{} than in \cN.'' Observed: $+0.072$, $+0.078$, $+0.083$.
\item \emph{Undecided.} ``In the public run the cost of an isotropic table makes more than half of its rise by 70k steps, like the share.'' Observed: $41\%$.
\item \emph{Undecided.} ``Relative to a 3{,}000-update network, the loss over all positions changes with age by at least $0.08$ nats less in \cT{} than in \cN{} (seed 0).'' Observed: $-0.046$. The same rule for the mean of three seeds was then met ($-0.118\pm0.037$; seeds 1 and 2: $-0.171$, $-0.139$).
\item \emph{Undecided.} ``At the best step of each network, an isotropic table is within $0.04$ nats at context positions in \cB{} and \cT, in both remaining seeds.'' Observed: $+0.000$ and $-0.076$ (seed 0: $+0.001$); the miss is on the side of \cB{} being faster.
\ExtraRules{}
\item \emph{A prediction outside the formal rules that failed.} ``Rare rows of a decayed table collapse to zero.'' Observed: only rows of tokens absent from the text do; rare rows keep $0.33$ of their size in \cN.
\item \emph{A check that is not used in this paper.} From the published schedule of another model with an exempt table \citep{donaghy2026open1b}, Eq.~\eqref{eq:walk} with $c$ between $0.84$ and $1.00$ gives a final table rms of $0.0723$ to $0.0853$; the rule required the measured value to fall in this band. Observed: $0.0697$ ($c=0.81$; median row $0.0754$, 99th percentile $0.0820$). Undecided by its rule.
\end{enumerate}
Several statements of earlier drafts of this account were withdrawn after checks of this kind, among them that the whole lag is a consequence of the takeover (the step term appears without it), that a single ratio governs all terms, that the trunk has lost no plasticity without qualification (with moment-matched tables old checkpoints are behind at context positions), and that the two mismatches describe relearning when the trunk is trained as well (\S\ref{sec:limits}).

\end{document}